\documentclass[9pt,journal,compsoc]{IEEEtran}
\usepackage[colorinlistoftodos, textwidth=4cm, shadow]{todonotes}

\usepackage[switch]{lineno}
\usepackage{subfigure}

\usepackage{tikz}
\usepackage{pdflscape}
\usepackage{afterpage}
\usepackage{pifont}
\usepackage{parskip}
\usepackage{authblk}
\usepackage{multirow}
\usepackage{colortbl}
\usepackage{tabularx}
\usepackage[
colorlinks=true,
linkcolor=blue,
urlcolor=blue,
citecolor=blue,
pdftex,
]{hyperref}
\usepackage[numbers]{natbib}
\usepackage{booktabs}
\usepackage{rotating}
\usepackage{graphicx}
\graphicspath{{Figures/}}
\usepackage{threeparttable}
\usepackage{url}

\usepackage{color}
\usepackage{todonotes}

\usepackage{comment} 

\RequirePackage[normalem]{ulem} 
\RequirePackage{color}\definecolor{RED}{rgb}{1,0,0}\definecolor{BLUE}{rgb}{0,0,1} 
\providecommand{\DIFaddbegin}{} 
\providecommand{\DIFaddend}{} 
\providecommand{\DIFdelbegin}{} 
\providecommand{\DIFdelend}{} 

\begin{document}
\title{The Evolution of First Person Vision Methods: \\ A Survey}

\author[1,2]{A. Betancourt\thanks{a.betancourt@tue.nl}}
\author[1]{P. Morerio\thanks{pietro.morerio@ginevra.dibe.unige.it}}
\author[1]{C.S. Regazzoni\thanks{carlo@dibe.unige.it}}
\author[2]{M. Rauterberg\DIFdelbegin 
\DIFdelend \DIFaddbegin \thanks{g.w.m.rauterberg@tue.nl \\ \\ \emph{This work was supported in part by the Erasmus Mundus joint Doctorate in Interactive and Cognitive Environments, which is funded by the EACEA, Agency of the European Commission under EMJD ICE.}}\DIFaddend }
\affil[1]{\footnotesize Department of Naval, Electric, Electronic and Telecommunications Engineering
DITEN - University of Genova, Italy}
\affil[2]{\footnotesize Designed Intelligence Group, Department of Industrial Design, Eindhoven University of Technology, Eindhoven, Netherlands}
\date{}
\maketitle

\begin{abstract}

The emergence of new wearable technologies such as action cameras and
smart-glasses has increased the interest of computer vision scientists in the
First Person perspective. Nowadays, this field is attracting attention  and
investments of companies aiming to develop commercial devices with First Person
Vision recording capabilities. Due to this interest, an increasing demand of
methods to process these videos, possibly in real-time, is expected. Current
approaches present a particular combinations of different image features and
quantitative methods to accomplish specific objectives like object detection,
activity recognition, user machine interaction and so on. This paper summarizes
the evolution of the state of the art in First Person Vision video analysis
between 1997 and 2014, highlighting, among others, most commonly used features,
methods, challenges and opportunities within the field.

\end{abstract}


\section{Introduction}

\begin{figure}
     \centering
     \includegraphics[width=0.45\textwidth]{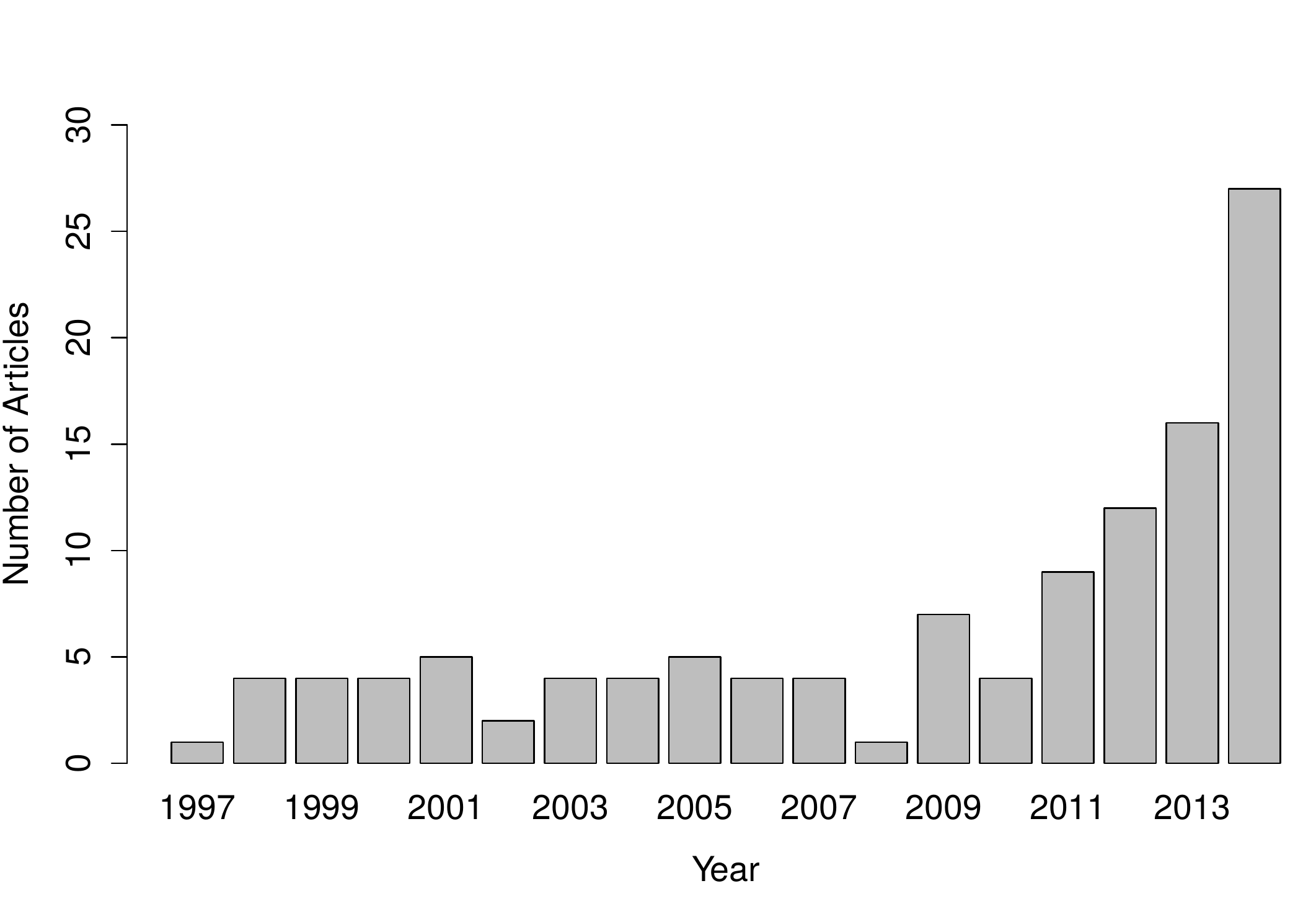}
     \caption{Number of articles per year directly related to FPV video analysis. This plot contains the articles published until 2014, to the best of our knowledge}
     \label{fig:numbArticles}
\end{figure}

Portable head-mounted cameras, able to record dynamic high quality first-person
videos, have become a common item among sportsmen over the last five years.
These devices represent the first commercial attempts to record experiences
from a first-person perspective. This technological trend is a follow-up of the
academic results obtained in the late 1990s, combined with the growing interest
of the people to record their daily activities. Up to now, no consensus has yet
been reached in literature with respect to naming this video perspective.
\emph{First Person Vision} (FPV) is arguably the most commonly used, but other
names, like \emph{Egocentric Vision} or \emph{Ego-vision} has also recently
grown in popularity.  The idea of recording and analyzing videos from this
perspective is not new in fact, several such devices have been developed for
research purposes over the last 15 years \cite{Mann1998, Mayol2000, Mayol2005a,
Hodges2006, Blum2006}. Figure \ref{fig:numbArticles} shows the growth in the
number of articles related to FPV video analysis between 1997 and 2014. Quite
remarkable is the seminal work carried out by the Media lab (MIT) in the late
1990s and early 2000s
\cite{Starner1998,Starner1998a,Schiele1999a,Schiele1999,Starner1999,Aoki1999},
and the multiple devices proposed by Steve Mann who, back in 1997
\cite{Mann1997}, described the field with these words : 

\begin{quotation} \emph{``Let's imagine a new approach to computing in which
the apparatus is always ready for use because it is worn like clothing. The
computer screen, which also serves as a viewfinder, is visible at all times and
performs multi-modal computing (text and images)''.} \end{quotation}

\begin{figure}[htp]
    \begin{center}
    \subfigure[Google glasses (U.S. Patent D659,741 - May 15, 2012).]{
    \includegraphics[width=0.42\linewidth]{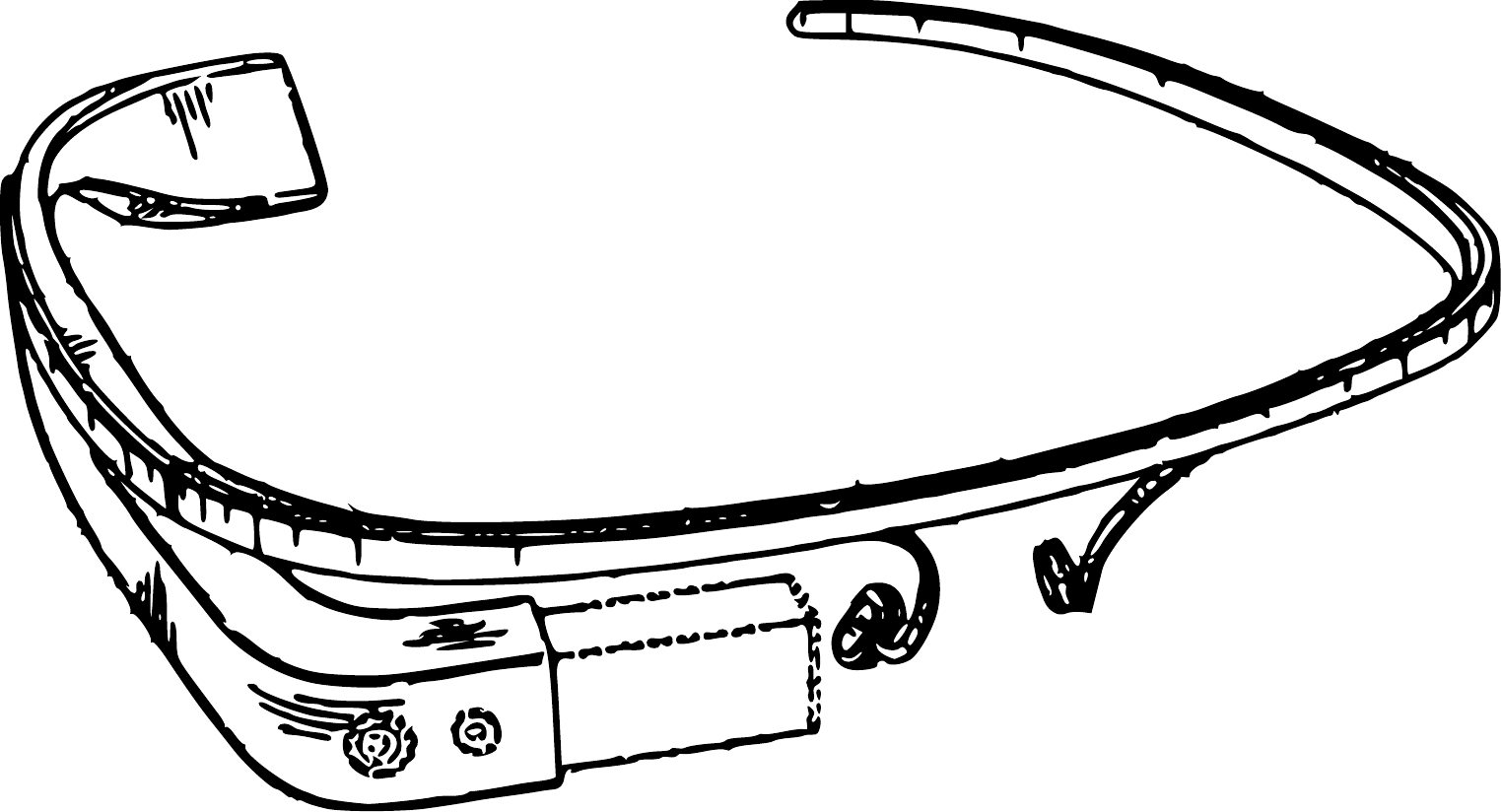} \label{fig:gglasses}
    }
    \quad
    \subfigure[Microsoft augmented reality glasses (U.S. Patent Application 20120293548 - Nov 22, 2012).]{
    \includegraphics[width=0.42\linewidth]{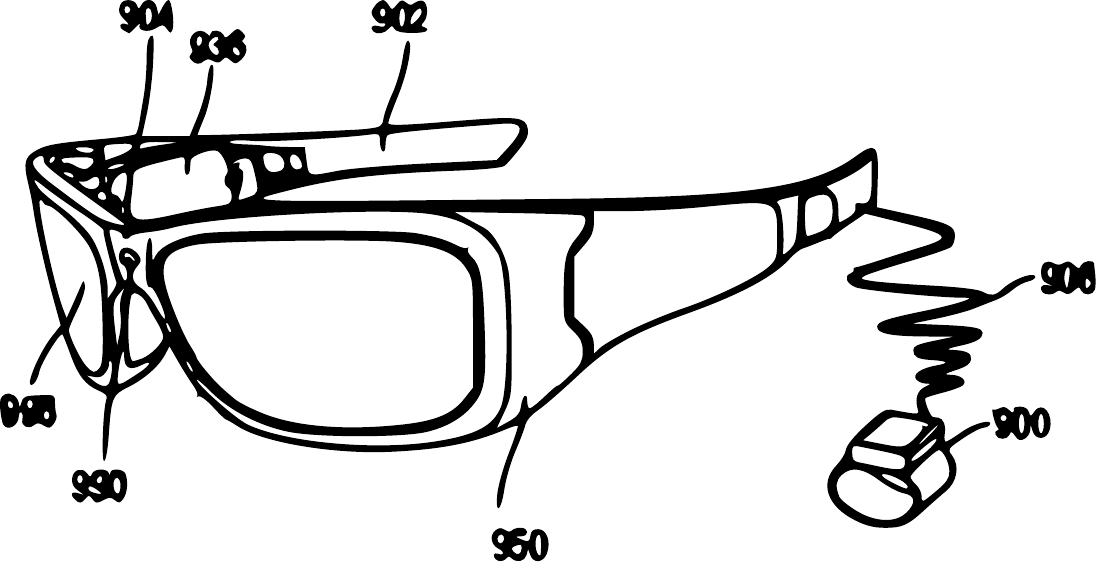} \label{fig:mglasses}
    }\\
    \caption{Examples of the commercial smart patents. (a) Google patent of the smart-glasses; (b) Microsoft patent of an augmented reality wearable device.}
    \end{center}
    \label{fig:sens_spec}
\end{figure}

Recently, in the awakening of this technological trend, several companies have
been showing interest in this kind of devices (mainly \emph{smart-glasses}),
and multiple patents have been presented. Figure  \ref{fig:sens_spec} shows the
devices patented in 2012 by Google and Microsoft. Together with its patent,
Google also announced \emph{Project Glass}, as a strategy to test its device
among a exploratory group of people. The project was introduced by showing
short previews of the Glasses' FPV recording capabilities, and its ability to
show relevant information to the user through the head-up display.

Remarkably, the impact of the Glass Project (wich the most significant attempt
to commercialize wearable technology up to date) is to be ascribed not only to
its hardware, but also to the appeal of its underlying operating system. The
latter continues to bring a large group of skilled developers, thus in turn
making a significant boost in the number of prospective applications for
smart-glasses, a phenomenon that has happened with smartphones several years
ago. On one hand, the range of application fields that could
benefit from smart-glasses is wide and applications are expected in areas like military
strategy, enterprise applications, tourist services \cite{Serra2013}, massive
surveillance \cite{Mann2003}, medicine \cite{Karaman2010}, driving assistance
\cite{Liu2014b}, among others. On the other hand, what was until now
considered as a consolidated research field, needs to be re-evaluated and restated under the light of this
technological trend: wearable technology and the first person perspective rise
important issues, such as privacy and battery life, in addition to new algorithmic
challenges \cite{Nguyen2009}. 

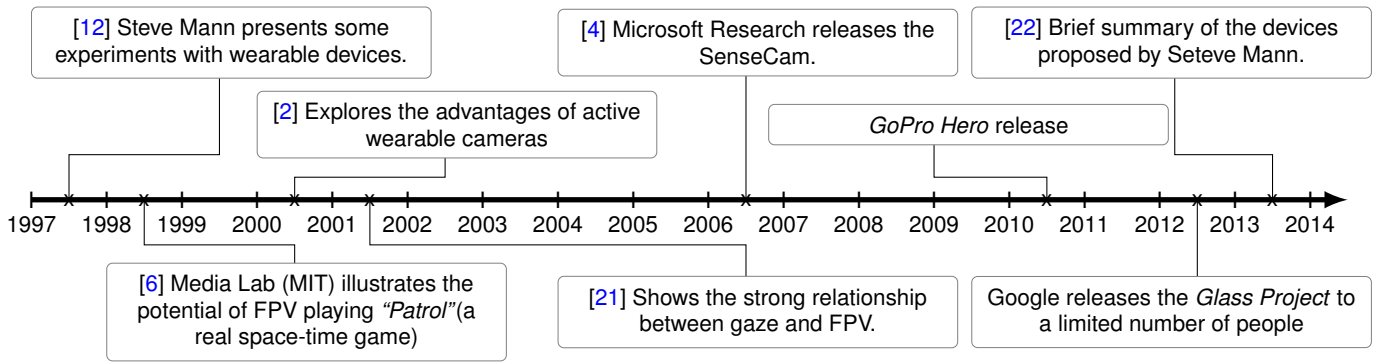
\begin{figure*}[htp]
\begin{center}
\begin{tikzpicture}[scale=1]
\small \sf 
\tikzset{l1/.style={draw=gray, ultra thin, rounded corners=.5ex, fill=white,text width=5cm, text badly centered,  inner sep=1ex, above = 1cm, anchor=west}}
\tikzset{l2/.style={draw=gray, ultra thin, rounded corners=.5ex, fill=white,text width=5cm, text badly centered,  inner sep=1ex, below = 5em, anchor=west}}
\tikzset{l3/.style={draw=gray, ultra thin, rounded corners=.5ex, fill=white,text width=5cm, text badly centered,  inner sep=1ex, above = 7em, anchor=west}}
\tikzset{l4/.style={draw=gray, ultra thin, rounded corners=.5ex, fill=white,text width=5cm, text badly centered,  inner sep=1ex, below = 11em, anchor=west}}
\tikzset{tick/.style={below=3pt}}
\tikzset{thinline/.style={ultra thin}}
\draw (0,0) -- (10,0);

\def\firsty{1997}
\def\lasty{2014}

\pgfmathparse{\lasty - \firsty}
\xdef\nsteps{\pgfmathresult}

\pgfmathparse{17/(\nsteps)}
\xdef\widthStep{\pgfmathresult}

\draw (0,0)[line width=2pt,->, -latex] -- (17.5,0);

\pgfmathparse{0*\widthStep + 0.5*\widthStep}
\xdef\x{\pgfmathresult}
\draw (\x, 0cm) node(link){x} (\x, 0cm) -- (\x,0.6cm) -- (\x+2,0.6cm)--  (\x+2, +2.3cm);
\draw (\x-0.5,0)   node(B1) [l3]    {\cite{Mann1997} Steve Mann presents some experiments with wearable devices.};

\pgfmathparse{1*\widthStep + 0.5*\widthStep}
\xdef\x{\pgfmathresult}
\draw (\x, 0cm) node(link){x} (\x, 0cm) -- (\x,-0.6cm) -- (\x+2,-0.6cm)--  (\x+2, -1.3cm);
\draw (\x-0.5,0)   node(B1) [l2]    {\cite{Starner1998} Media Lab (MIT) illustrates the potential of FPV playing \emph{``Patrol''}(a real space-time game)};

\pgfmathparse{3*\widthStep + 0.5*\widthStep}
\xdef\x{\pgfmathresult}
\draw (\x, 0cm) node(link){x} (\x, 0cm) -- (\x,0.3cm) -- (\x+2,0.3cm) -- (\x+2, +1.3cm);
\draw (\x-0.5,0)   node(B1) [l1]    {\cite{Mayol2000} Explores the advantages of active wearable cameras};

\pgfmathparse{4*\widthStep + 0.5*\widthStep}
\xdef\x{\pgfmathresult}
\draw (\x, 0cm) node(link){x} (\x, 0cm) -- (\x,-0.6cm) -- (\x+5,-0.6cm)--  (\x+5, -1.3cm);
\draw (\x+2.5,0)   node(B1) [l2]    {\cite{Land2001} Shows the strong relationship between gaze and FPV.};

\pgfmathparse{9*\widthStep + 0.5*\widthStep}
\xdef\x{\pgfmathresult}
\draw (\x, 0cm) node(link){x} (\x, 0cm) -- (\x, +2.3cm);
\draw (\x-2.5,0)   node(B1) [l3]    {\cite{Hodges2006} Microsoft Research releases the SenseCam.};

\pgfmathparse{13*\widthStep + 0.5*\widthStep}
\xdef\x{\pgfmathresult}
\draw (\x, 0cm) node(link){x} (\x, 0cm) -- (\x,0.3cm) -- (\x-1.5,0.3cm)--  (\x-1.5, 0.9cm);
\draw (\x-3.7,0)   node(B1) [l1]    {\emph{GoPro Hero} release};

\pgfmathparse{15*\widthStep + 0.5*\widthStep}
\xdef\x{\pgfmathresult}
\draw (\x, 0cm) node(link){x} (\x, 0cm)  -- (\x, -1.3cm);
\draw (\x-3,0)   node(B1) [l2]    {Google releases the \emph{Glass Project} to a limited number of people};

\pgfmathparse{16*\widthStep + 0.5*\widthStep}
\xdef\x{\pgfmathresult}
\draw (\x, 0cm) node(link){x} (\x, 0cm) -- (\x,0.6cm) -- (\x-1.3,0.6cm)--  (\x-1.3, +2.3cm);
\draw (\x-4,0)   node(B1) [l3]    {\cite{Mann2013} Brief summary of the devices proposed by Seteve Mann.};


\def\year{\firsty}

\foreach \y in {0,...,\nsteps}
{
    \draw (\y*\widthStep, 0)  node(A1) [tick]  {\year}; 
    \draw (\y*\widthStep, 0) node (\y) {I};

    \pgfmathparse{int(\year+1)}
    \xdef\year{\pgfmathresult}
};

\end{tikzpicture}
\caption{Some of the more important works and commercial announcements in FPV.} \label{tl:general}
\end{center}
\end{figure*}

This paper summarizes the state of the art in FPV video analysis and its
temporal evolution between 1997 and 2014, analyzing the challenges and
opportunities of this video perspective. It reviews the main characteristics of
previous studies using tables of references, and the main events and relevant
works using timelines. As an example, Figure \ref{tl:general} presents some of
the most important papers and commercial announcements in the general evolution
of FPV.  We direct interested readers to the \emph{must read} papers presented
in this timeline. In the following sections, more detailed timelines are
presented according to the objective addressed in the summarized papers.
{\color{black} The categories and conceptual groups presented in this survey
reflects our schematic perception of the field coming from a detailed study of
the existent literature. We are confident that the proposed categories are wide
enough to conceptualize existent methods, however due to the growing speed of
the field they could require future updates}. As will be shown in the coming
sections, the strategies used during the last $20$ years are very
heterogeneous. Therefore, rather than provide a comparative structure between
existing methods and features, the objective of this paper is to highlight
common points of interest and relevant future lines of research.  The
bibliography presented in this paper is mainly in FPV. However, some particular
works in classic video analysis are also mentioned to support the analysis. The
latter are cited using italic font as a visual cue. 

To the best of our knowledge, the only paper summarizing the general ideas of
the FPV is \cite{Kanade2012}, which presents a wearable device and several
possible applications.  Other related reviews include the following:
\textit{\cite{Guan2011}} reviews the activity recognition methods with multiple
sensors; \cite{Doherty2013} analyzes the use of wearable cameras for medical
applications; \cite{Mayol2005a} presents some challenges of an active wearable
device.

In the remainder of this paper, we summarize existent methods in FPV, according
to a hierarchical structure we propose, highlighting the more relevant works
and the temporal evolution of the field. Section \ref{sec:hierar} introduces
general characteristics of FPV and the hierarchical structure, which is later
used to summarize the current methods according to their final objective, the
subtasks performed and the features used. In section \ref{sec:datasets} we
briefly present the publicly-available FPV datasets. Finally, section
\ref{sec:future} discusses some future challenges and research opportunities in
this field.

\section{First Person Vision (FPV) video analysis} \label{sec:hierar}

During the late 1990s and early 2000s, the advances in FPV analysis were mainly
performed using highly elaborated devices, typically proprietarily developed by
different research groups. The list of devices proposed is wide, where each
device was usually presented in conjunction with their potential applications
and a large array of sensors which only envy from modern devices in their
design, size and commercial capabilities. The column ``Hardware'' in Table
\ref{tab:sumtable} summarizes these devices. The remaining columns of this
table are explained in section \ref{sec:objectives}. Nowadays, current devices
could be considered as the embodiment of the futuristic perspective of the
already mentioned pioneering studies. Table \ref{tab:commercial} shows the
currently available commercial projects and their embedded sensors. Such
devices are grouped in three categories:

\begin{itemize}

    \item{\emph{\bf{Smart-glasses:}} Smart-glasses have multiple sensors,
    processing capabilities and a head-up display, making them ideal to develop
    real time methods and to improve the interaction between the user and its
    device. Besides, smart-glasses are nowadays seen as the starting point of
    an augmented reality system. However, they cannot be considered a mature
    product until major challenges, such as battery life, price and target
    market, are solved. The future of these devices is promising, but it is
    still not clear if they will be adopted by the users on a daily basis like
    smartphones, or whether they will become specialized task-oriented devices
    like industrial glasses, smart-helmets, sport devices, etc.}

    \item{\emph{\bf{Action cameras:}} commonly used by sportsmen and
    lifeloggers. However, the research community has been using them as a tool
    to develop methods and algorithms while anticipating the commercial
    availability of the smart-glasses during the coming years. Action cameras
    are becoming cheaper, and are starting to exhibit (albeit still somewhat
    limited) processing capabilities.}

    \item{\emph{\bf{Eye trackers:}} have been successfully applied to analyze
    consumer behaviors in commercial environments. Prototypes are available
    mainly for research purposes, where multiple applications have been proposed
    in conjunction with FPV. {\color{black}Despite the potential of these
    devices, their popularity is highly affected by the price of their components
    and the obtrusiveness of the eye tracker sensors, which is commonly carried
    out using an eye pointing camera.}}

\end{itemize}

\begin{table*}
\scriptsize
\caption{Commercial approaches to wearable devices with FPV video recording capabilities}
\centering
\begin{threeparttable}[b]
\begin{tabular}{r|ccccccccccccc}
\toprule
      & \begin{sideways}Camera\end{sideways} & \begin{sideways}Eye Tracking\end{sideways} & \begin{sideways}{Microphone}\end{sideways} & \begin{sideways}GPS\end{sideways} & \begin{sideways}Accelerometer\end{sideways} & \begin{sideways}Gyroscope\end{sideways} & \begin{sideways}Magnetometer\end{sideways} & \begin{sideways}Altitude\end{sideways} & \begin{sideways}Light Sensor\end{sideways} & \begin{sideways}Proximity Sensor\end{sideways} & \begin{sideways}Body-Heat Detector\end{sideways} & \begin{sideways}Temperature Sensor\end{sideways} & \begin{sideways}Head-Up Display\end{sideways} \\
\midrule
\arrayrulecolor{gray}
Google Glasses           & \ding{51} &          & \ding{51} & \ding{51} & \ding{51} & \ding{51} & \ding{51} &           & \ding{51} & \ding{51} &           &           & \ding{51} \\
Epson Moverio            & \ding{51} &          & \ding{51} & \ding{51} & \ding{51} & \ding{51} & \ding{51} &           &           &           &           &           & \ding{51} \\
Recon Jet                & \ding{51} &          & \ding{51} & \ding{51} & \ding{51} & \ding{51} & \ding{51} &           &           &           &           & \ding{51} & \ding{51} \\
Vuzix M100               & \ding{51} &          & \ding{51} &           & \ding{51} & \ding{51} & \ding{51} &           & \ding{51} & \ding{51} &           &           & \ding{51} \\
GlassUp                  & \ding{51} &          & \ding{51} &           & \ding{51} & \ding{51} & \ding{51} &           & \ding{51} &           &           &           & \ding{51} \\
Meta                     & \ding{51} &          & \ding{51} & \ding{51} & \ding{51} & \ding{51} &           &           &           &           &           &           & \ding{51} \\
Optinvent Ora-s          & \ding{51} &          & \ding{51} & \ding{51} & \ding{51} & \ding{51} & \ding{51} &           & \ding{51} &           &           &           & \ding{51} \\
SenseCam                 & \ding{51} &          & \ding{51} &           & \ding{51} &           &           & \ding{51} & \ding{51} &           & \ding{51} & \ding{51} &           \\
Lumus                    & \ding{51} &          & \ding{51} &           & \ding{51} & \ding{51} & \ding{51} &           &           &           &           &           & \ding{51} \\ \hline
Pivothead                & \ding{51} &          & \ding{51} &           &           &           &           &           &           &           &           &           &           \\
GoPro                    & \ding{51} &          & \ding{51} &           &           &           &           &           & \ding{51} &           &           &           &           \\
Looxcie camera           & \ding{51} &          & \ding{51} &           &           &           &           &           &           &           &           &           &           \\ 
Epiphany Eyewear         & \ding{51} &          & \ding{51} &           &           &           &           &           &           &           &           &           &           \\\hline
SMI Eye tracking Glasses & \ding{51} &\ding{51} & \ding{51} &           &           &           &           &           &           &           &           &           &           \\
Tobii                    & \ding{51} &\ding{51} & \ding{51} &           &           &           &           &           &           &           &           &           &           \\
\arrayrulecolor{black}
\bottomrule
\end{tabular}%

  \label{tab:commercial}
  \begin{tablenotes}
      \item [1] Other projects such as \emph{Orcam}, \emph{Nissan}, \emph{Telepathy}, \emph{Olympus MEG4.0}, \emph{Oculon} and \emph{Atheer} have been officially announced by their producers but no technical specifications have been already presented.
      \item [2] According to unofficial online sources, other companies like \emph{Apple}, \emph{Samsung}, \emph{Sony}, \emph{Oakley} could be working on their own versions of similar devices, however no information has been officially announced up to date. \emph{Microsoft} recently announced the Hololens but not technical specifications have been officially presented. 
      \item [2] This data is created on January 2015.
      \item [3] In \cite{Kanade2012} one multi-sensor device is presented for research purposes.
      \end{tablenotes}
\end{threeparttable}\\
\end{table*}%

FPV video analysis gives some methodological and practical advantages, but also
inherently brings a set of challenges that need to be addressed
\cite{Kanade2012}. On one hand, FPV solves some problems of the classical video
analysis and offers extra information:

\begin{itemize}

    \item \emph{Videos of the main part of the scene:} Wearable devices allow
    the user to (even unknowingly) record the most relevant parts of the scene
    for the analysis, thus reducing the necessity for complex controlled
        multi-camera systems \cite{Fathi2012}.      

    \item \emph{Variability of the datasets:} Due to the increasing commercial
    interest of the technology companies, a large number of FPV videos is
    expected in the future, making it possible for the researchers to obtain
    large datasets that differ among themselves significantly, as discussed in
    section \ref{sec:datasets}. 

    \item \emph{Illumination and scene configuration:} Changes in the
    illumination and global scene characteristics could be used as an important
    feature to detect the scene in which the user is involved, e.g. detecting
    changes in the place where the activity is taking place, as in \cite{Lu2013}. 

    \item \emph{Internal state inference:} According to \cite{Yarbus1967}, eye
    and head movements are directly influenced by the person's emotional state.
    As already done with smartphones \textit{\cite{bisio2013opportunistic}},
    this fact can be exploited to infer the user's emotional state, and provide
    services accordingly.

    \item \emph{Object positions:} Because users tend to see the objects while
    interacting with them, it is possible to take advantage of the prior
    knowledge of the hands' and objects' positions, e.g. active objects tend to
    be closer to the center, whereas hands tend to appear in the bottom left and
    bottom right part of the frames \cite{Philipose2009, Betancourt2014a}.

\end{itemize}

On the other hand, FPV itself also presents multiple challenges, which
particularly affect the choice of the features to be extracted by low level
processing modules (feature selection is discussed in details in section
\ref{sec:features}):

\begin{itemize}

    \item \emph{Non static cameras:} One of the main characteristics of FPV
    videos is that cameras are always in movement. This fact makes it difficult
    to differentiate between the background and the foreground
    \cite{Ghosh2012}. Camera calibration is not possible and often scale,
    rotation and/or translation-invariant features are required in higher level
    modules. 

    \item \emph{Illumination conditions:} The locations of the videos are
    highly variable and uncontrollable (e.g. visiting a touristic place during
    a sunny day, driving a car at night, brewing coffee in the kitchen). This
    makes it necessary to deploy robust methods for dealing with the
    variability in illumination. Here\, shape descriptors may be preferred to
    color-based features \cite{Betancourt2014a}.

    \item \emph{Real time requirements:} One of the motivations for FPV video
    analysis is its potential of being used for real time activities. This
    implies the need for the real time processing capabilities
    \cite{Morerio2013}. 

    \item \emph{Video processing:} Due to the embedded processing capabilities
    (for smart-glasses), it is important to define efficient computational
    strategies to optimize battery life, processing power and communication
    limits among the processing units. At this point, cloud computing could be
    seen as the most promising candidate tool to turn the FPV video analysis
    into an applicable framework for daily use. However, a real time cloud
    processing strategy requires further development in video compressing
    methods and communication protocols between the device and the cloud
    processing units.

\end{itemize}

\begin{figure}
    \centering
    \includegraphics[width=0.48\textwidth]{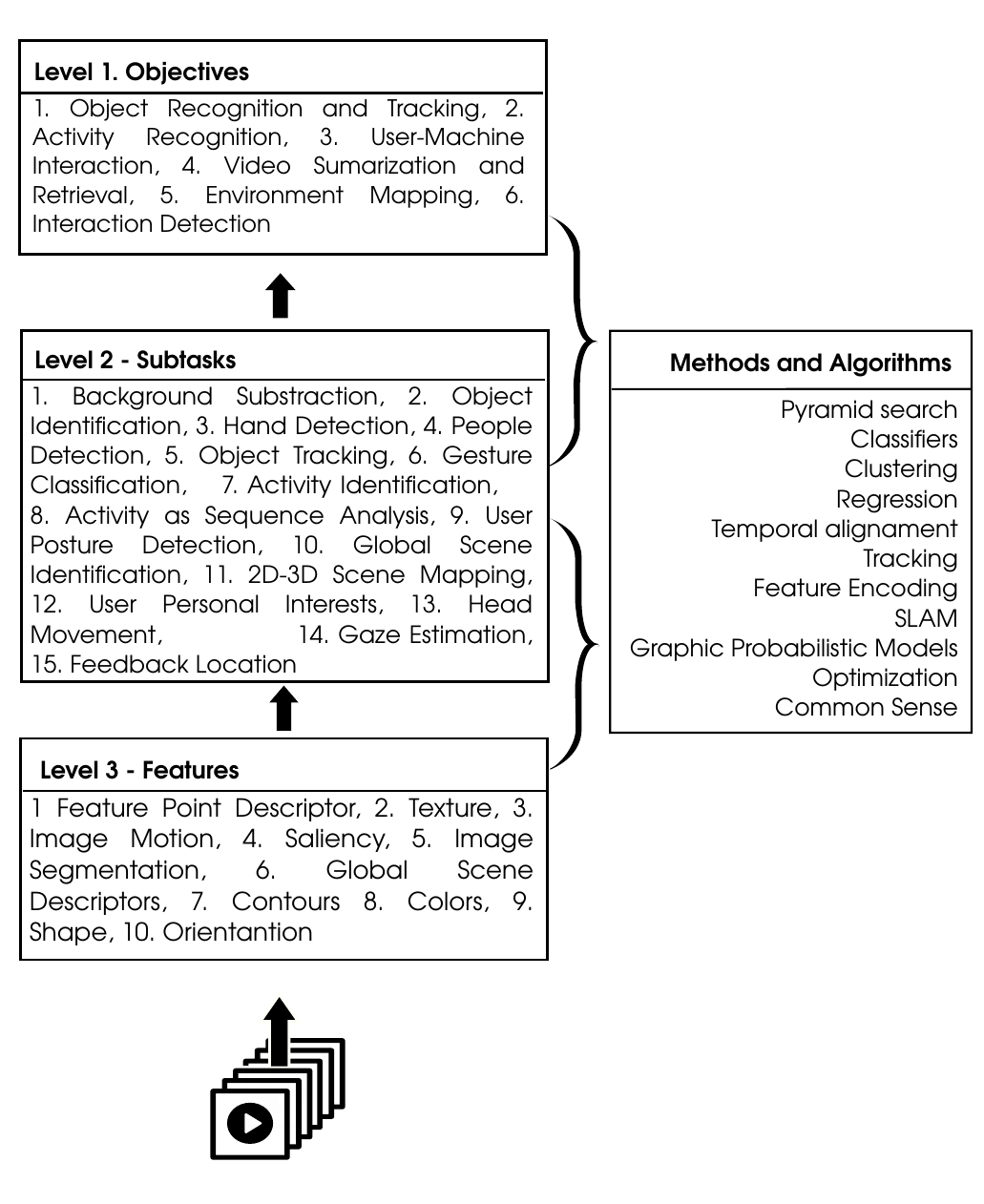}
    \caption{Hierarchical structure to explain the state of the art in FPV video analysis.}
    \label{fig:hierarchical}
\end{figure}

The rest of this chapter summarizes FPV video analysis methods according to a
hierarchical structure, as shown in Figure \ref{fig:hierarchical}, starting
from the raw video sequence (bottom) to the desired objectives (top). Section
\ref{sec:objectives} summarizes the existent approaches according to 6 general
objectives (Level 1). Section \ref{sec:subtasks} divides these objectives in 15
weakly dependent subtasks (Level 2). Section briefly introduces the most
commonly used image features, presenting their advantages and disadvantages,
and relating them with objectives. Finally, section \ref{sec:methalg}
summarizes the quantitative and computational tools used to process data,
moving from one level to the other.  In our literature review, we found that
existing  methods are commonly presented as combinations of the aforementioned
levels. However, no standard structure is presented,  making it difficult for
other researchers to replicate existing methods or improve the state of the
art. We propose this hierarchical structure as an attempt to cope with this
issue.

\subsection{Objectives} \label{sec:objectives}

Table \ref{tab:sumtable} summarizes a total of 117 articles. The articles are
divided in six objectives according to the main goal addressed in each of them.
The left side of the table contains the six objectives described in this
section, and on the right side, extra groups related to hardware, software,
related surveys and conceptual articles, are given. The category named
''Particular Subtasks`` is used for articles focused on one of the subtasks
presented in section \ref{sec:subtasks}. The last column shows the positive
trend in the number of articles per year, and is plotted in Figure
\ref{fig:numbArticles}.

\begin{table*}
\scriptsize
\centering
\caption{Summary of the articles reviewed in FPV video analysis according to the main objective}
\begin{threeparttable}[b]
\begin{tabular}{r|rrrrrr|rrrrr|r}
\toprule
  \multicolumn{1}{r}{}    & \multicolumn{6}{c}{Objective}                 & \multicolumn{6}{c}{Extra Categories} \\
\midrule
\multicolumn{1}{c|}{\begin{sideways}\textbf{Year}\end{sideways}} & \begin{sideways}\textbf{Object Recognition and Tracking}\end{sideways} & \begin{sideways}\textbf{Activity Recognition}\end{sideways} & \begin{sideways}\textbf{User-Machine Interaction}\end{sideways} & \begin{sideways}\textbf{Video Summarization and Retrieval}\end{sideways} & \begin{sideways}\textbf{Environment Mapping}\end{sideways} & \begin{sideways}\textbf{Interaction Detection}\end{sideways} & \begin{sideways}\textbf{Particular Subtasks}\end{sideways} & \begin{sideways}\textbf{Related Software Design}\end{sideways} & \begin{sideways}\textbf{Hardware}\end{sideways} & \begin{sideways}\textbf{Conceptual Academic Articles}\end{sideways} & \begin{sideways}\textbf{Related Surveys}\end{sideways} & \begin{sideways}\textbf{\# Articles Reviewed}\end{sideways} \\
\midrule
1997&&&&&&&&&\cite{Mann1997}&&&1\\
1998&\cite{Starner1998a,Starner1998,Bradski1998}&\cite{Starner1998}&&&&&&&\cite{Mann1998}&&&4\\
1999&\cite{Schiele1999a,Clarkson1999}&\cite{Schiele1999a,Aoki1999}&\cite{Schiele1999}&&&&&&\cite{Schiele1999}&&&4\\
2000&\cite{Farringdon2000}&\cite{Clarkson2000}&\cite{Kurata2000}&&&&&&\cite{Mayol2000}&&&4\\
2001&\cite{Kurata2001a}&&\cite{Kurata2001,Kojima2001}&\cite{Aizawa2001}&&&\cite{Land2001}&&&&&5\\
2002&&&&\cite{Sawahata2002,Gemmell2002}&&&&\cite{Gemmell2002}&&&&2\\
2003&&&\cite{DeVaul2003}&\cite{Sawahata2003,Hori2003}&&&&&&\cite{Mann2003}&&4\\
2004&&&\cite{Bane2004,Kolsch2004a}&\cite{Gemmell2004}&&&&\cite{Gemmell2004}&\cite{Mann2004}&&&4\\
2005&\cite{Mayol2005,Sun2009}&\cite{Mayol2005}&\cite{Tenmoku2005}&\cite{Tancharoen2005,Aizawa2005}&&&&\cite{,Aizawa2005}&\cite{Mayol2005a}&&&5\\
2006&&\cite{Blum2006}&\cite{Bane2006,Kolsch2006}&&&&&&\cite{Hodges2006,Blum2006}&&&4\\
2007&\cite{Castle2007,Castle2007a}&\cite{Wu2007}&&&\cite{Castle2007,Davison2007,Castle2007a}&&&&&&&4\\
2008&\cite{Castle2008}&&&&\cite{Castle2008}&&&&&&&1\\
2009&\cite{Philipose2009,Sun2009}&\cite{Yi2009,Spriggs2009,Sundaram2009}&\cite{Makita2009}&&&&&&&\cite{Nguyen2009}&&7\\
2010&\cite{Ren2010}&\cite{Karaman2010,Dovgalecs2010}&&\cite{Dovgalecs2010,Byrne2010}&&&&&&&&4\\
2011&\cite{Hebert2011,Fathi2011a}&\cite{Doherty2011,Fathi2011}&&\cite{Aghazadeh2011,Doherty2011,Kitani2011}&&&\cite{Yamada2011a}&&\cite{MichaelS.DevyverAkihiroTsukada2011}&&\cite{Guan2011}&9\\
2012&&\cite{Borji2012,Pirsiavash2012,Fathi2012a,Ghosh2012,Kitani2012,Ogaki2012}&\cite{Grasset2012}&&\cite{Park2012}&\cite{Fathi2012}&\cite{Yamada2012,Boujut2012}&&\cite{Kanade2012}&&\cite{Kanade2012}&12\\
2013&\cite{Li2013b,Li2013a,Morerio2013,Wang2013,Zariffa2013}&\cite{Li2013,GonzalezDiaz2013,Fathi2013}&\cite{Serra2013,Zhang2013}&\cite{Lu2013,Fathi2013}&&\cite{Ryoo2013}&\cite{Martinez2013}&&\cite{Mann2013}&\cite{Starner2013}&\cite{Doherty2013}&16\\
2014  & * & \cite{Narayan2014, Matsuo2014, Damen2014, Poleg2014, Zheng, Zhan2014} &       & ** &       & \cite{Alletto2014,Alletto} & \cite{Betancourt2014a, Lee2014} & \cite{Han2014} &       & \cite{Scheck2014} &       & 27 \\
\bottomrule
\end{tabular}
\label{tab:sumtable}
  \begin{tablenotes}
      \item [*\ ] \cite{Wang2014, Liu2014a, Feng2014, Poleg2014a, Rogez2014, Rogez2014a, Templeman2014, Buso2014, Liu2014b}
      \item [**] \cite{Xiong2014, Min2014, Bolanos2014, Barker2014, Okamoto2014, Arev2014}
      \end{tablenotes}
\end{threeparttable}\\
\end{table*}%


Note from the table that the most commonly explored objective is \emph{Object
Recognition and Tracking}. We identify it as the base of more advanced
objectives such as \emph{Activity Recognition}, \emph{Video Summarization and
Retrieval} and \emph{Environment Mapping}. Another often studied objective is
\emph{User-Machine Interaction} because of its potential in Augmented Reality.
Finally, a recent research line denoted as \emph{Interaction Detection} allows
the devices to infer situations in which the user is involved. Along with this
section, we present some details of how existent methods have addressed each of
these 6 objectives. One important aspect is that some methods use multiple
sensors within a data-fusion framework. For each objective, several examples of
data-fusion and multi-sensor approaches are mentioned.

\subsubsection{Object recognition and tracking}\label{sec:objectrecognitionandtracking}

\begin{figure*}[htp]
\begin{center}
\begin{tikzpicture}[scale=1]
\small \sf 
\tikzset{l1/.style={draw=gray, ultra thin, rounded corners=.5ex, fill=white,text width=5cm, text badly centered,  inner sep=1ex, above = 1cm, anchor=west}}
\tikzset{l2/.style={draw=gray, ultra thin, rounded corners=.5ex, fill=white,text width=5cm, text badly centered,  inner sep=1ex, below = 5em, anchor=west}}
\tikzset{l3/.style={draw=gray, ultra thin, rounded corners=.5ex, fill=white,text width=5cm, text badly centered,  inner sep=1ex, above = 7em, anchor=west}}
\tikzset{l4/.style={draw=gray, ultra thin, rounded corners=.5ex, fill=white,text width=5cm, text badly centered,  inner sep=1ex, below = 11em, anchor=west}}
\tikzset{tick/.style={below=3pt}}
\tikzset{thinline/.style={ultra thin}}
\draw (0,0) -- (10,0);

\def\firsty{1997}
\def\lasty{2014}

\pgfmathparse{\lasty - \firsty}
\xdef\nsteps{\pgfmathresult}

\pgfmathparse{17/(\nsteps)}
\xdef\widthStep{\pgfmathresult}

\draw (0,0)[line width=2pt,->, -latex] -- (17.5,0);

\pgfmathparse{2*\widthStep + 0.5*\widthStep}
\xdef\x{\pgfmathresult}
\draw (\x, 0cm) node(link){x} (\x, 0cm) --  (\x, 1.3cm);
\draw (\x-2.5,0)   node(B1) [l1]    {\cite{Jones1999} developes a pixel by pixel classifier to locate human skin in videos.};

\pgfmathparse{3*\widthStep + 0.5*\widthStep}
\xdef\x{\pgfmathresult}
\draw (\x, 0cm) node(link){x} (\x, 0cm) -- (\x,-0.3cm) -- (\x,-0.3cm)--  (\x, -1.3cm);
\draw (\x-2.5,0)   node(B1) [l2]    {\cite{Farringdon2000} presents the ``Augmented Memory'', following the ideas presented in \cite{Schiele1999}, as a killer application in the field of FPV};

\pgfmathparse{8*\widthStep + 0.5*\widthStep}
\xdef\x{\pgfmathresult}
\draw (\x, 0cm) node(link){x} (\x, 0cm) -- (\x,+0.6cm) -- (\x,+0.6cm)--  (\x, +1.3cm);
\draw (\x-2.5,0)   node(B1) [l1]    {\cite{Mayol2005} proposes the first dataset for object recognition in FPV.};

\pgfmathparse{12*\widthStep + 0.5*\widthStep}
\xdef\x{\pgfmathresult}
\draw (\x, 0cm) node(link){x} (\x, 0cm) -- (\x,-0.6cm) -- (\x-2.8,-0.6cm)--  (\x-2.8, -1.3cm);
\draw (\x-5.5,0)   node(B1) [l2]    {\cite{Philipose2009} proposes a challenging dataset of 42 daily use objects recorded at different scales and perspectives.};

\pgfmathparse{14*\widthStep + 0.5*\widthStep}
\xdef\x{\pgfmathresult}
\draw (\x, 0cm) node(link){x} (\x, 0cm) --  (\x, +1.3cm);
\draw (\x-2.5,0)   node(B1) [l1]    {\cite{Fathi2011a} uses Multiple Instance Learning to reduce the labeling requirements in object recognition.};

\pgfmathparse{16*\widthStep + 0.5*\widthStep}
\xdef\x{\pgfmathresult}
\draw (\x, 0cm) node(link){x} (\x, 0cm) -- (\x,-0.6cm) -- (\x-1.3,-0.6cm)--  (\x-1.3,-1.3cm);
\draw (\x-4,0)   node(B1) [l2]    {\cite{Li2013b} train a pool of models to deal with changes in illumination.};


\def\year{\firsty}

\foreach \y in {0,...,\nsteps}
{
    \draw (\y*\widthStep, 0)  node(A1) [tick]  {\year}; 
    \draw (\y*\widthStep, 0) node (\y) {I};

    \pgfmathparse{int(\year+1)}
    \xdef\year{\pgfmathresult}
};

\end{tikzpicture}
\caption{Some of the more important works in \emph{object recognition and tracking.}\label{tl:object}}
\end{center}
\end{figure*}
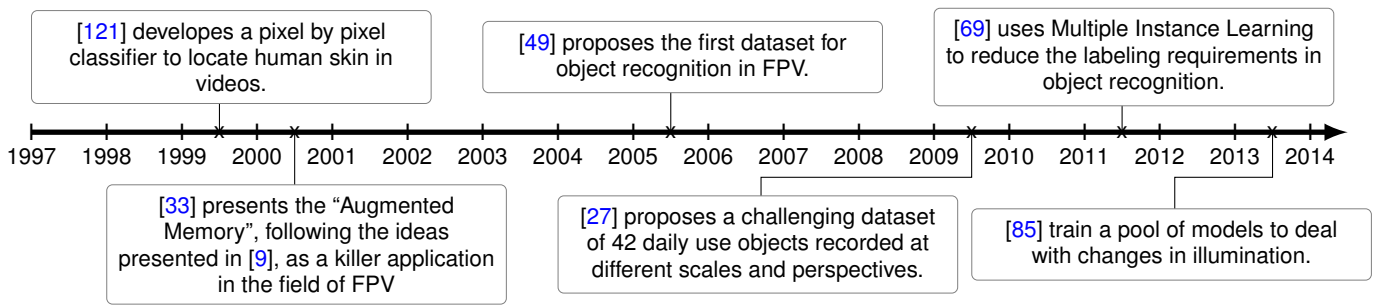

\emph{Object recognition and tracking} is the most explored objective in FPV,
and its results are commonly used as a starting point for more advanced tasks,
such as activity recognition. Figure \ref{tl:object} summarizes some of the
most important papers that focused on this objective. 

In addition to the general opportunities and challenges of the FPV perspective,
this objective introduces important aspects to be considered: i) Because of the
uncontrolled characteristics of the videos, the number of objects,  as well as
their type, scale and point of view, are unknown  \cite{Philipose2009,
Pirsiavash2012}. ii) Active objects, as well as user's hands, are frequently
occluded. iii) Because of the mobile nature of the wearable cameras, it is not
easy to create background-foreground models. iv) The camera location makes it 
possible to build a priori information about the objects' position
\cite{Philipose2009, Betancourt2014a}. 

Hands are among the most common objects in the user's field of view, and a
proper detection, localization, and tracking could be a main input for other
objectives. The authors in \cite{Betancourt2014a} highlight the difference
between hand-detection and hand-segmentation, particularly in the framework of
wearable devices where the number of deployed computational resources, directly
influences the battery life of the devices. {\color{black}In general, due to
the hardware availability and price, hand-detection and tracking is usually
carried out  using RGB videos. However, \cite{Rogez2014,Rogez2014a} uses a
chest-mounted RGB-D camera to improve the hand-detection and tracking
performance in realistic scenarios.} According to \cite{Mayol2005}, hand
detection could be divided into model-driven and data-driven methods. 

Model-driven methods search for the best matching configuration of a
computational hand model (2D or 3D) to recreate the image that is being shown
in the video \cite{Schlattmann2007,Schlattman2007, Rehg1994, Sun2009, Rogez2014,Rogez2014a}. These
methods are able to infer detailed information of the hands, such as the
posture, but in exchange large computational resources, highly controlled
environments or extra sensors (e.g. Depth Cameras) could be required.


Data-driven methods use image features to detect and segment users' hands. The
most commonly used features for this purpose are the color histograms looking
to exploit the particular chromaticism of human skin, especially in suitable
color spaces like HSV and YCbCr \cite{Morerio2013, Serra2013, Li2013b,
Li2013a}. Color-based methods can be considered as the evolution of the
pixel-by-pixel skin classifiers proposed in \textit{\cite{Jones1999}}, in which
color histograms are used to decide whether a pixel represents human skin.
Despite their advantages, the color-based methods are far from being an optimal
solution. Two of their more important restrictions are: i) The computational
cost, because in each frame they have to solve the $O(n^2)$ problem implied by
the pixel-by-pixel classification. ii) Their results highly influenced by
significant changes in illumination, for example indoor and outdoor
videos\cite{Betancourt2014a}. To reduce the computational cost, some authors
suggest the use of superpixels \cite{Serra2013, Morerio2013, Li2013a},
however, an exhaustive comparison of the computational times of both approaches
is still pending, and computationally efficient superpixel methods applied to
video (especially FPV video) are still at an early stage
\cite{morerio2014optimizing}. Regarding the noisy results, the authors in
\cite{Li2013b, Serra2013} train a pool of models and automatically select the
most appropriate depending on the current environmental conditions.

In addition to hands, there is an uncountable number of objects that could
appear in front of the user, whose proper identification could lead to
development of some of the most promising applications of FPV. An example is ``The
Virtual Augmented Memory(VAM)'' proposed by \cite{Farringdon2000}, where the
device is able to identify objects, and to subsequently relate them to previous
information, experiences or common knowledge available online. {\color{black}An
interesting extension of the VAM is presented in \cite{Bettadapura}, where the
user is spatially located using his video, and is shown relevant information
about the place or a particular event. In the same line of research, recent
approaches have been trying to fuse information from multiple wearable cameras
to recognize when the users are being recorded by a third person without
permission. This is accomplished in \cite{Poleg2014a,Hoshen2014} using the
motion of the wearable camera as the identity signature, which is subsequently
matched in the third person videos without disclosing private information such as the
face or the identity of the user.} 


The augmented memory is not the only application of object recognition. The
authors in \cite{Pirsiavash2012} develop an \emph{activity recognition method}
which based only a list of the used objects in the recorded video . Despite the
importance of these applications, the problem of recognition is far from being
solved due to the large amount of objects to be identified as well as the
multiple positions and scales from which they could be observed. It is here
that machine learning starts playing a key role in the field, offering tools to
reduce the required knowledge about the objects \cite{Fathi2011a} or exploiting
web services (such as Amazon Turk) and automatic mining for training purposes
\cite{Spain2010, Ghosh2012, Berg2006, Wu2007}.   

Once the objects are detected, it is possible to track their movements. In the
case of the hands, some authors use the coordinates of center as the reference
point \cite{Morerio2013}, while others go a step further and use dynamic models
\cite{Kolsch2004a, Kolsch2006}. Dynamic models are widely studied and are
successfully used to track hands, external objects \cite{Davison2007,
Castle2007, Davison2007, Castle2008, Castle2007a}, or faces of other people
\cite{Bradski1998}.

\subsubsection{Activity recognition}\label{subsec:activity_recognition}

\begin{figure*}[htp]
\begin{center}
\begin{tikzpicture}[scale=1]
\small \sf 
\tikzset{l1/.style={draw=gray, ultra thin, rounded corners=.5ex, fill=white,text width=5cm, text badly centered,  inner sep=1ex, above = 1cm, anchor=west}}
\tikzset{l2/.style={draw=gray, ultra thin, rounded corners=.5ex, fill=white,text width=5cm, text badly centered,  inner sep=1ex, below = 5em, anchor=west}}
\tikzset{l3/.style={draw=gray, ultra thin, rounded corners=.5ex, fill=white,text width=5cm, text badly centered,  inner sep=1ex, above = 3.8em, anchor=west}}
\tikzset{l4/.style={draw=gray, ultra thin, rounded corners=.5ex, fill=white,text width=5cm, text badly centered,  inner sep=1ex, below = 11em, anchor=west}}
\tikzset{tick/.style={below=3pt}}
\tikzset{thinline/.style={ultra thin}}
\draw (0,0) -- (10,0);

\def\firsty{1997}
\def\lasty{2014}

\pgfmathparse{\lasty - \firsty}
\xdef\nsteps{\pgfmathresult}

\pgfmathparse{17/(\nsteps)}
\xdef\widthStep{\pgfmathresult}

\draw (0,0)[line width=2pt,->, -latex] -- (17.5,0);

\pgfmathparse{2*\widthStep + 0.5*\widthStep}
\xdef\x{\pgfmathresult}
\draw (\x, 0cm) node(link){x} (\x, 0cm) -- (\x,+0.3cm) -- (\x,+0.3cm)--  (\x, +1.3cm);
\draw (\x-2.5,0)   node(B1) [l1]    {\cite{Starner1998} uses markov models to detect the user action in the  game \emph{``patrol''}.};

\pgfmathparse{5*\widthStep + 0.5*\widthStep}
\xdef\x{\pgfmathresult}
\draw (\x, 0cm) node(link){x} (\x, 0cm) -- (\x,-0.6cm) -- (\x,-0.6cm) -- (\x, -1.3cm);
\draw (\x-2.5,0)   node(B1) [l2]    {\cite{Yu2002} shows the advantages of eye-trackers for activity recognition.};

\pgfmathparse{9*\widthStep + 0.5*\widthStep}
\xdef\x{\pgfmathresult}
\draw (\x, 0cm) node(link){x} (\x, 0cm) -- (\x,+0.3cm) -- (\x-1,+0.3cm)--  (\x-1, +1.3cm);
\draw (\x-3.5,0)   node(B1) [l3]    {\cite{Hodges2006} presents the ``SenseCam'' a multi-sensor device subsequently used for activity recognition.};

\pgfmathparse{12*\widthStep + 0.5*\widthStep}
\xdef\x{\pgfmathresult}
\draw (\x, 0cm) node(link){x} (\x, 0cm) -- (\x,0.3cm) -- (\x+1.5,0.3cm)--  (\x+1.5, 1.3cm);
\draw (\x-1,0)   node(B1) [l1]    {\cite{Sundaram2009} models activities as sequences of events using only FPV videos};

\pgfmathparse{14*\widthStep + 0.5*\widthStep}
\xdef\x{\pgfmathresult}
\draw (\x, 0cm) node(link){x} (\x, 0cm) -- (\x,-0.6cm) -- (\x,-0.6cm)--  (\x, -1.3cm);
\draw (\x-2.5,0) node(B1) [l2]    {\cite{Pirsiavash2012} summarizes the importance of object detection in activity recognition, dealing with different perspectives and scales.};


\def\year{\firsty}

\foreach \y in {0,...,\nsteps}
{
    \draw (\y*\widthStep, 0)  node(A1) [tick]  {\year}; 
    \draw (\y*\widthStep, 0) node (\y) {I};

    \pgfmathparse{int(\year+1)}
    \xdef\year{\pgfmathresult}
};

\end{tikzpicture}
\caption{Some of the more important works in activity recognition.} \label{tl:activityRecognition}

\end{center}
\end{figure*}
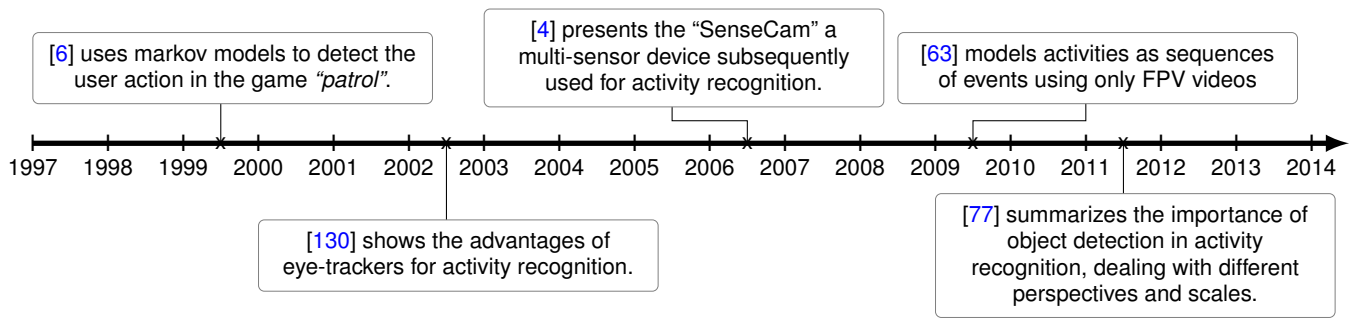

An intuitive step in the hierarchy of objectives is \emph{Activity
Recognition}, aimedat identifying what the user is doing in a particular video
sequence. Figure \ref{tl:activityRecognition} presents some of the most
relevant papers on this topic. A common approach in activity recognition is to
consider an activity as a sequence of events that can be modeled as Markov
Chains or as Dynamic Bayesian Networks (DBNs) \cite{Starner1998,
Schiele1999a,Clarkson2000, Blum2006, Sundaram2009}. Despite the promising
results of this approach, the main challenge to be solved is the scalability to
multiple user and multiple strategies to solve a similar task. 

Recently, two major methodological approaches for \emph{activity recognition}
are becoming popular: object based and motion based recognition. Object based
methods aim to infer the activity using the objects appearing in video sequence
\cite{Sundaram2009, Fathi2011, Pirsiavash2012}, assuming of course that the
activities can be described by the required group of objects( e.g. prepare a
cup of coffee requires coffee, water and a spoon). This approach opens the door
to highly scalable strategies based on web mining to know the objects usually
required for different activities. However, after all, this approach depends on
a proper \emph{Object Recognition} step and on its own challenges (Section
\ref{sec:objectrecognitionandtracking}). {\color{black}Following an alternative
path, during the last 3 years, some authors have been using the fact that
different kind of activities create different body motions and as consequence
different motion patterns in the video, for example: walking, running, jumping,
skiing, reading, watching movies, among others \cite{Kitani2011, Ogaki2012,
Poleg2014}. It is remarkable the discriminative power of motion features for
this kind of activities and the robustness to deal with the illumination and
the color skin challenges.}

\emph{Activity recognition} is one of the fields that has drawn most benefits
from the use of multiple sensors. This strategy started growing in popularity
with the seminal work of Clarkson et al. \cite{Clarkson2000,Clarkson1999} where
basic activities are identified using FPV video jointly with audio signals. An
intuitive realization of the multi-sensor strategy allows to reduce the
dependency between \emph{Activity Recognition} and \emph{Object Recognition},
by using Radio-Frequency Identification (RFID) tags in the objects
\cite{Wu2007, Krahnstoever2005, Patterson2005, Philipose2004}. However, the use
of RFIDs reduces the applicability to environments previously tagged. The list
of multiple sensors does not end with audio and RFIDs, it also contains
Inertial Measurement Units \textit{\cite{Spriggs2009}}, multiple sensors of the
``SenseCam\footnote{Wearable device developed by Microsoft Research in
Cambridge with accelerometers, thermometer, infrared and light sensor}''
\cite{Doherty2011,Byrne2010}, GPS \citep{Ghosh2012}, and eye-trackers
\cite{Yi2009, Fathi2012a, Yamada2012,Yamada2011a,Li2013}.

\subsubsection{User-machine interaction}

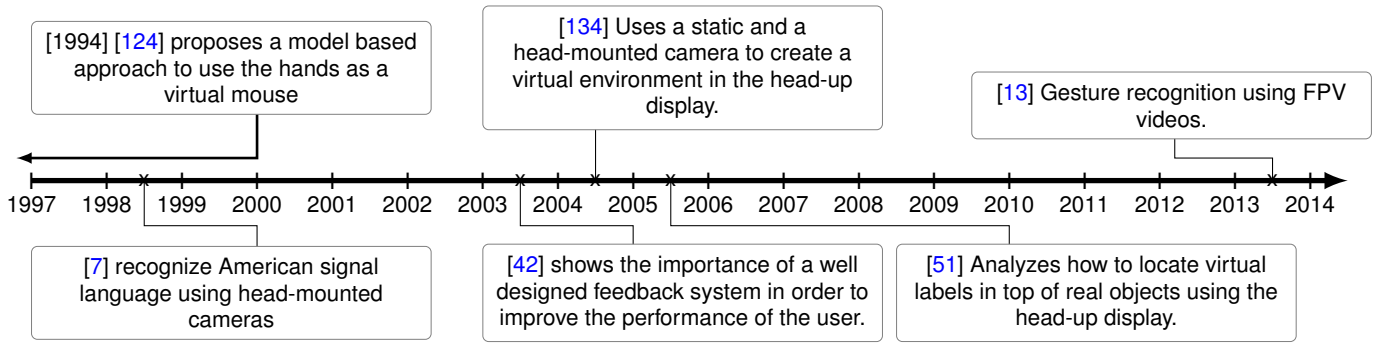
\begin{figure*}[htp]
\begin{center}
\begin{tikzpicture}[scale=1]
\small \sf 
\tikzset{l1/.style={draw=gray, ultra thin, rounded corners=.5ex, fill=white,text width=5cm, text badly centered,  inner sep=1ex, above = 1cm, anchor=west}}
\tikzset{l2/.style={draw=gray, ultra thin, rounded corners=.5ex, fill=white,text width=5cm, text badly centered,  inner sep=1ex, below = 5em, anchor=west}}
\tikzset{l3/.style={draw=gray, ultra thin, rounded corners=.5ex, fill=white,text width=5cm, text badly centered,  inner sep=1ex, above = 1.5cm, anchor=west}}
\tikzset{l4/.style={draw=gray, ultra thin, rounded corners=.5ex, fill=white,text width=5cm, text badly centered,  inner sep=1ex, below = 2cm, anchor=west}}
\tikzset{tick/.style={below=3pt}}
\tikzset{thinline/.style={ultra thin}}
\draw (0,0) -- (10,0);

\def\firsty{1997}
\def\lasty{2014}

\pgfmathparse{\lasty - \firsty}
\xdef\nsteps{\pgfmathresult}

\pgfmathparse{17/(\nsteps)}
\xdef\widthStep{\pgfmathresult}

\draw (0,0)[line width=2pt,->, -latex] -- (17.5,0);

\pgfmathparse{0*\widthStep + 0.5*\widthStep}
\xdef\x{\pgfmathresult}
\draw (\x+2.5, +1.3cm) -- (\x+2.5,0.3cm)[line width=1pt,style={<-}, -latex] -- (\x-0.7, 0.3cm) ;
\draw (\x-0.5,0)   node(B1) [l3]    {[1994] \cite{Rehg1994} proposes a model based approach to use the hands as a virtual mouse};

\pgfmathparse{1*\widthStep + 0.5*\widthStep}
\xdef\x{\pgfmathresult}
\draw (\x, 0cm) node(link){x} (\x, 0cm) -- (\x,-0.6cm) -- (\x+1.5,-0.6cm)--  (\x+1.5, -1.3cm);
\draw (\x-1.5,0)   node(B1) [l2]    {\cite{Starner1998a} recognize American signal language using head-mounted cameras};

\pgfmathparse{6*\widthStep + 0.5*\widthStep}
\xdef\x{\pgfmathresult}
\draw (\x, 0cm) node(link){x} (\x, 0cm) -- (\x,-0.6cm) -- (\x+1.5,-0.6cm)--  (\x+1.5, -1.3cm);
\draw (\x-0.5,0)   node(B1) [l2]    {\cite{DeVaul2003} shows the importance of a well designed feedback system in order to improve the performance of the user.};

\pgfmathparse{7*\widthStep + 0.5*\widthStep}
\xdef\x{\pgfmathresult}
\draw (\x, 0cm) node(link){x} (\x, 0cm) -- (\x, +1.3cm);
\draw (\x-1.5,0)   node(B1) [l3]    {\cite{Kolsch2004} Uses a static and a head-mounted camera to create a virtual environment in the head-up display.};

\pgfmathparse{8*\widthStep + 0.5*\widthStep}
\xdef\x{\pgfmathresult}
\draw (\x, 0cm) node(link){x} (\x, 0cm) -- (\x,-0.6cm) -- (\x+4.5,-0.6cm)--  (\x+4.5, -1.3cm);
\draw (\x+3,0)   node(B1) [l2]    {\cite{Tenmoku2005} Analyzes how to locate virtual labels in top of real objects using the head-up display.};

\pgfmathparse{16*\widthStep + 0.5*\widthStep}
\xdef\x{\pgfmathresult}
\draw (\x, 0cm) node(link){x} (\x, 0cm) -- (\x,0.3cm) -- (\x-1.3,0.3cm)--  (\x-1.3, +1.3cm);
\draw (\x-4,0)   node(B1) [l1]    {\cite{Serra2013} Gesture recognition using FPV videos.};


\def\year{\firsty}

\foreach \y in {0,...,\nsteps}
{
    \draw (\y*\widthStep, 0)  node(A1) [tick]  {\year}; 
    \draw (\y*\widthStep, 0) node (\y) {I};

    \pgfmathparse{int(\year+1)}
    \xdef\year{\pgfmathresult}
};

\end{tikzpicture}
\caption{Some of the more important works and commercial announcements in FPV.}\label{tl:usermachineinteraction}
\end{center}
\end{figure*}

As already mentioned, smart-glasses open the door to new ways for interaction
between the user and his device. The device, being able to give feedback to the
user, allows to close the interaction loop originated by the visual information
captured and interpreted by the camera. Due to the scope of this paper, only
approaches related to FPV video analysis are presented (we omit other sensors,
such as audio and touch panels), categorizing them based on two approaches: i)
the user sends information to the device, and ii) the device uses the
information of the video to show the feedback to the user.  Figure
\ref{tl:usermachineinteraction} shows some of the most important works
concerning \emph{User-machine interaction}.

In general, the interaction between the user and his device starts with
intentional or unintentional command. An \emph{intentional command} is a signal
sent by the user using his hands through his camera. This kind of interaction
is not a recent idea and several approaches have been proposed, particularly
using static cameras \textit{\cite{rivag2005, garcia2011}}, which, as mentioned
in section \ref{sec:objectrecognitionandtracking}, can not be straightforwardly
applied to FPV due to the mobile nature of wearable cameras. A traditional
approach is to emulate the mouse of computers with the hands
\cite{Rehg1994,Kurata2000, Kurata2001}, allowing the user to point and click at
virtual objects created in the head-up display. Other approaches look for more
intuitive and technology focused ways of interaction. For example, the authors
in \cite{Serra2013} develop a gesture recognition algorithm to be used in an
interactive museum using 5 different gestures: ``point out'', ``like'',
``dislike'', ``OK'' and ``victory''. In \cite{Zhang2013}, the head movements of
the user are used to assist a robot in the task of finding a hidden object in
a controlled environment. Under this perspective some authors combine static
and wearable cameras\cite{Kolsch2004, Starner1998a}. Quite remarkable are the
results of Starner in 1998, being able to recognize American signal language
with an efficiency of 98\% with a static camera and head mounted camera. As is
evident, hand-tracking methods can give important cues in this objective
\cite{Morshidi2014,Spruyt2013,Shan2007,Kolsch2004a}, and make it possible to
use features such as position, speed or acceleration of the users' hands.

\emph{Unintentional commands} are triggers activated by the device using
information about the user without his conscious intervention, for example: i)
the user is cooking by following a particular recipe (Activity Recognition),
and the device could monitor the time of different steps without the user
previously asking for it. ii) The user is looking at a particular item [Object
Recognition] in a store [GPS or Scene Recognition] then the device could show
price comparisons and reviews. \emph{Unintentional commands} could be detected
using the results of other FPV objectives, the measurements of its sensors, or
behavioural routines learned from the user while previously using his device,
among others. From our point of view, these kinds of commands could be the next
step of user-machine interaction for smart-glasses, and a main enabler to
reduce the required time to interact with the device \cite{Starner2013}.

Regarding the second part of the interaction loop, it is important to properly
design the feedback system to know when, where, how, and which information
should be shown to the user. In order to accomplish this, several issues must
be considered in order to avoid misbehaviour of the system that could work
against the user's performance in addressing relevant tasks \cite{DeVaul2003}.
In this line, multiple studies develop methods to optimally locate virtual
labels in the user's visual field, without occluding the important parts of the
scene \cite{Makita2009,Tenmoku2005,Grasset2012}.

\subsubsection{Video summarization and retrieval}


The main task of \emph{Video summarization and retrieval} is to create tools to
explore and visualize the most important parts of large FPV video sequences
\cite{Lu2013}. The objective and main issue is perfectly summarized in
\cite{Aizawa2001} with the following sentence: \emph{``We want to record our
entire life by video. However, the problem is how to handle such a huge
data''}. In general, existing methods define importance functions to select the
more relevant subsequences or frames of the video, and later cut or accelerate
the less important ones \cite{Okamoto2014}. Recent studies define the
importance function using the objects appearing in the video \cite{Ghosh2012},
their temporal relationships and causalities \cite{Lu2013}, {\color{black}or as
a similarity function, in terms of its composition, between them and
intentional pictures taken with a traditional cameras \cite{Xiong2014}. A
remarkable result is achieved in \cite{Kitani2011, Poleg2014} using motion
features to segment videos according to the activity performed by the user.
This work is a good example of how to take advantage of the camera movements in
FPV, usually considered as a challenge, to achieve good classification rates}. 


The use of multiple sensors is common within this objective, and remarkable
fusions have been made using brain measurements in \cite{Aizawa2001,
Sawahata2002}, gyroscopes, accelerometers, GPS, weather information and skin
temperature in \cite{Sawahata2003,Hori2003,Tancharoen2005}, and online
available pictures in \cite{Xiong2014}. An alternative approach to video
summarization is presented in \cite{Park2012} and \cite{Arev2014}, where
multiple FPV videos of the same scene are unified using the collective
attention of the wearable cameras as an importance function.  {\color{black}In
order to define whether the two videos recorded from different cameras are
pointing at the same scene, the authors in \cite{Ben-Artzi2014} use superpixels
and motion features to propose a similarity measurement}. Finally, it is
significant to mention that ``Video summarization and retrieval'' has led to
important improvements in the design of the databases and visualization methods
to store and explore the recorded videos \cite{Gemmell2002,Gemmell2004}. In
particular, this kind of developments can be considered an important tool for
reducing computational requirements in the devices, as well as alleviate
privacy issues related with the place where videos are stored.

\subsubsection{Environment Mapping}

\emph{Environment Mapping} aims at the construction of a 2D or 3D virtual
representation of the environment surrounding the user. In general, the of
variables to be mapped can be divided in two categories: physical variables,
such as walls and object locations, and intangible variables, such as attention
points. Physical mapping is the more explored of the two groups. It started to
grow in popularity with \citep{Davison2007}, which showed how, by using
multiple sensors, Kalman Filters and monoSLAM, it is possible to elaborate a
virtual map of the environment. Subsequently, this method was improved by
adding object identification and location as a preliminary stage
\cite{Castle2007, Castle2007a}. Physical mapping is one of the more complex
tasks in FPV, particularly when 3D maps are required due to the calibration
restrictions.  This problem can be partially alleviated by using a multi-camera
approach to infer the depth \cite{Castle2008, Kanade2012}.  Research on
\emph{intangible variables}, can be considered an emerging field in FPV.
Existent approaches define attention points and attraction fields, mapping them
in rooms with multiple people interacting \cite{Arev2014}. 

\subsubsection{Interaction detection}

The objectives described above are mainly focused on the user of the device as
the only person that matters in the scene. However, they hardly take into
account the general situation in which the user is involved. We label the group
of methods aiming to recognize the types of interaction that the user is having
with other people as \emph{Interaction Detection}.  One of the main purposes in
this objective is \emph{social interaction detection}, as proposed by
\cite{Fathi2012}.  In their paper, the authors inferred the gaze of the other
people and used it to recognize human interactions as monologues, discussions
or dialogues.  Another approach in this field was proposed by \cite{Ryoo2013},
which detected different behaviors of the people surrounding the user (e.g.
hugging, punching, throwing objects, among others). Despite not being widely
explored yet, this objective can be considered one of the most promising and
innovative ones a in FPV due to the mobility and personalization capabilities
of the coming devices.


\subsection{Subtasks} \label{sec:subtasks}

As explained before, the proposed structure is based on objectives which are
highly co-dependent. Moreover, it is common to find that the output of one
objective is subsequently used as the input for the other (e.g.  activity
recognition usually depends on object recognition). For this reason, a common
practice is to first address small subtasks, and later merge them to accomplish
main objectives. Based on the literature review, we propose a total of 15
subtasks.  Table \ref{tab:objvssub} shows the number of articles analyzed in
this survey that use a subtask (columns) in order to address a particular
objective (rows).  It is important to highlight the many-to-many relationship
among objectives and subtasks, which means that a subtask could be used to
address different objectives, and one objective could require multiple
subtasks. To mention some: i) hand detection, as a subtask, could be the
objective itself in object recognition, \cite{Morerio2013}, but could also give
important cues in activity recognition \cite{Fathi2012a}; moreover, it could be
the main input in the user-machine interaction \cite{Serra2013}. ii) The
authors in \citep{Pirsiavash2012} performed object recognition to subsequently
infer the performed activity. As we reckon that their names are
self-explanatory, we omit separate explanation of each of the subtasks, with
the possible exceptions of the following: i) \emph{Activity as a Sequence}
analyzes an activity as a set of ordered steps; ii) \emph{2D-3D Scene Mapping}
builds a 2D or 3D virtual representation of the scene recorded; iii) \emph{User
Personal Interests} identifies the parts in the video sequence potentially
interesting for the user using physiological signals such as
brainwaves\cite{Sawahata2002}; iv) \emph{Feedback location} identifies the
optimal place in the head-up display to locate the virtual feedback without
interfering with the user's visual field.

\begin{table*}
    \caption{Number of times that a subtask is performed to accomplish a specific objective}
  \centering
\begin{tabular}{r|ccccccccccccccc}
\toprule
  Objective    & \begin{sideways}Background Subtraction\end{sideways} & \begin{sideways}Object Identification\end{sideways} & \begin{sideways}Hand Detection and Segmentation\end{sideways} & \begin{sideways}People Detection\end{sideways} & \begin{sideways}Object Tracking\end{sideways} & \begin{sideways}Gesture Identification\end{sideways} & \begin{sideways}Activity Identification\end{sideways} & \begin{sideways}Activity as Sequence Analysis\end{sideways} & \begin{sideways}User Posture Detection\end{sideways} & \begin{sideways}Global Scene Identification\end{sideways} & \begin{sideways}2D-3D Scene Mapping\end{sideways} & \begin{sideways}User Personal Interests\end{sideways} & \begin{sideways}Head Movement\end{sideways} & \begin{sideways}Gaze Estimation\end{sideways} & \begin{sideways}Feedback Location\end{sideways} \\
\midrule
Object Recognition and Tracking & 4     & 15    & 13    & 3     & 10    &       &       &       &       & 2     &       &       & 1     & 1     &  \\
Activity Recognition          &  3     & 8     & 3     & 1     &       &       & 13    & 2     & 1     & 8     & 1     &       & 6     & 6     &  \\  
User-Machine Interaction           &   &       & 6     &       & 3     & 2     &       &       &       &       &       &       & 1     &       & 3 \\
Video Summarization            & 1     & 4     & 1     & 4     &       &       & 5     & 1     &       & 4     &       & 1     & 2     & 1     &  \\
Environment Mapping                 &  & 3     &       &       & 4     &       &       &       &       &       & 5     &       &       & 1     &  \\
Interaction Detection               &  &       &       & 2     & 1     &       & 2     &       &       &       & 1     &       & 2     & 2     &  \\
\bottomrule
\end{tabular}%
  \label{tab:objvssub}%
\end{table*}%

As can be deduced from table \ref{tab:objvssub}, \emph{Hand detection} plays an
important role as the base for advanced objectives such as \emph{Object
Recognition} and \emph{User-Machine interaction}. \emph{Global scene
identification}, as well as \emph{Object Identification}, stand out as two
important subtasks for activity recognition. More in detail, the tight bound
between the \emph{Activity Recognition} and the \emph{Object Recognition}
supports the idea of \cite{Pirsiavash2012}, which states that Activity
Recognition is ``all about objects''. Moreover, the use of gaze estimation in
multiple objectives confirms the advantages of the recent trend of using
eye-trackers in conjunction with FPV videos. Finally, it can be noted that
\emph{Background Subtraction} has lost some of its reputation if compared with
fixed camera scenarios, due to the highly unstable nature of the backgrounds
when observed from the First-person perspective.

\subsection{Video and image features}\label{sec:features}

As mentioned before, FPV implies highly dynamic changes in the attributes and
characteristics of the scene. Due to these changes, an appropriate selection of
the features becomes critical in order to alleviate the challenges and exploit
the advantages presented in section \ref{sec:hierar}. As is well known, feature
selection is not a trivial task, and usually implies an exhaustive search in
the literature and extensive testing to identify which method leads to optimal
results. 

The process of feature extraction is carried out at different levels, starting
from the pixel level, with color channels of the image, and subsequently
extracting more elaborated indicators at the frame level, such as
\emph{saliency}, \emph{texture}, \emph{superpixels}, \emph{gradients}, etc. As
expected, these features can be used to address some of the subtasks, such as
object recognition or scene identification.  However, they do not include any
kind of dynamic information. To add dynamic information in the analysis,
different approaches can be followed, for example analyzing the geometrical
transformation between two frames to obtain image \emph{Motion} features such
as \emph{optical flow}, or aggregating frame level features in temporal
windows.  Usually, dynamic features tend to be computationally expensive, and
are therefore usually applied to objectives in which the video is processed
once the activities have finished.  Particularly interesting is the method
presented in \cite{morerio2014optimizing}, which uses the information of the
superpixels of the previous frame to initialize and compute the current frame
superpixels, thus reducing the computational complexity of the algorithm by
$60\%$. 

Table \ref{tab:subtvf} shows the most commonly used features in FPV to address
a particular subtask. The features are listed in the rows and the subtasks in
the columns. Note that \emph{color histograms} are by far the most commonly
used feature for almost all the subtasks, despite being highly criticized due
to their dependence on illumination changes. {\color{black}Another group of
features frequently used for several subtasks is \emph{Image Motion}.
Some of its most remarkable results are for \emph{Activity Recognition} in
\cite{Kitani2011, Poleg2014}, for \emph{Video Summarization} in
\cite{Okamoto2014}, and recently as the input of a Convolutional Neural Network
(CNN) to create a biometric sensor that is able to identify the user recording
the video in \cite{Hoshen2014}}. The use of \emph{Feature Point Descriptors}
(FPD) is also worth noting. As expected, they are popular for object
identification, but it is also remarkable their application to identify
relevant places such as touristic hotspots \cite{Aghazadeh2011, Karaman2010,
Dovgalecs2010}. Note from the table that the ``dynamic objectives'' like
\emph{Activity Recognition} and \emph{Video Summarization} are the ones which
take the most advantage of the \emph{Motion} features, while \emph{Object
Recognition} is mainly based on frame features such as \emph{FPD} and
\emph{Color histograms}.   


\begin{table*}
\scriptsize
\centering
    \caption{Number of times that each feature is used in to solve an objective or subtask}
\begin{tabular}{r|l|cccccc|ccccccccccccccc}
\toprule
      &       & \multicolumn{6}{c|}{Objectives}                & \multicolumn{15}{c}{Subtasks} \\
\midrule
      &       &       &       &       &       &       &       &       &       &       &       &       &       &       &       &       &       &       &       &       &       &  \\
      &       & \begin{sideways}Object Recognition and Tracking\end{sideways} & \begin{sideways}Activity Recognition\end{sideways} & \begin{sideways}User Machine interaction\end{sideways} & \begin{sideways}Video Summarization and Retrieval\end{sideways} & \begin{sideways}Environment Mapping\end{sideways} & \begin{sideways}Interaction Detection\end{sideways} & \begin{sideways}Background Substraction\end{sideways} & \begin{sideways}Object Identification\end{sideways} & \begin{sideways}Hand Detection and segementation\end{sideways} & \begin{sideways}People detection\end{sideways} & \begin{sideways}Object Tracking\end{sideways} & \begin{sideways}Gesture Clasification\end{sideways} & \begin{sideways}Activity Identification\end{sideways} & \begin{sideways}Activity as Sequence Analysis\end{sideways} & \begin{sideways}User Posture Detection\end{sideways} & \begin{sideways}Global Scene Identification\end{sideways} & \begin{sideways}2D-3D mapping\end{sideways} & \begin{sideways}User Personal Interests\end{sideways} & \begin{sideways}Head Movement\end{sideways} & \begin{sideways}Gaze estimation\end{sideways} & \begin{sideways}Feedback Location\end{sideways} \\ \midrule
\multirow{8}[2]{*}{FPD}                & SIFT                & 9     & 5     & 1     & 4     & 3     &       & 2     & 14    &       & 1     &       &       & 1     &       &       & 2     &       &       &       &       &  \\
                                       & GFTT                & 1     & 1     &       &       &       &       &       &       &       &       & 1     &       &       &       &       &       &       &       &       &       &  \\
                                       & BRIEF               & 2     &       &       &       &       &       &       &       & 1     &       &       &       &       &       &       & 1     &       &       &       &       &  \\
                                       & FAST                & 1     &       &       &       &       &       &       &       &       &       &       &       &       &       &       & 1     &       &       &       &       &  \\
                                       & SURF                & 2     & 6     &       & 1     &       &       &       & 2     &       & 1     & 1     &       & 2     &       &       & 2     &       &       &       &       &  \\
                                       & Diff. of Gaussians  &       &       &       & 1     &       &       &       & 1     &       &       &       &       &       &       &       &       &       &       &       &       &  \\
                                       & ORB                 & 1     &       &       &       &       &       &       &       & 1     &       &       &       &       &       &       &       &       &       &       &       &  \\
                                       & STIP                &       & 2     &       &       &       &       & 1     & 1     &       &       &       &       &       &       &       &       &       &       &       &       &  \\
\midrule
\multirow{5}[2]{*}{Texture}            & Wiccest             &       &       &       & 1     &       &       &       &       &       &       &       &       & 1     &       &       &       &       &       &       &       &  \\
                                       & Laplacian Transform & 1     &       &       &       &       &       &       &       & 1     &       &       &       &       &       &       &       &       &       &       &       &  \\
                                       & Edge Histogram      &       & 1     & 1     & 1     &       &       &       & 1     &       &       &       &       &       &       &       &       &       & 1     &       &       & 1\\
                                       & Wavlets             & 1     &       &       &       &       &       &       &       &       & 1     &       &       &       &       &       &       &       &       &       &       &  \\
                                       & Other               &       & 1     &       & 1     &       &       &       &       &       &       &       &       & 1     &       &       &       &       &       &       &       &  \\
\midrule
\multirow{3}[2]{*}{Saliency}           & GBVS                & 1     & 1     &       &       &       &       &       &       &       &       &       &       &       &       &       &       &       &       &       & 4     &  \\
                                       & Other               &       & 1     &       &       &       &       &       & 1     &       &       &       &       &       &       &       &       &       &       &       & 1     &  \\
                                       & MSSS                &       &       & 1     &       &       &       &       &       &       &       &       &       &       &       &       &       &       &       &       &       & 1\\
\midrule
\multirow{3}[2]{*}{Motion}             & Optical Flow        & 5     & 14    & 2     & 5     &       & 1     & 5     & 1     & 2     &       & 2     &       & 6     &       &       & 1     &       &       & 4     & 5     &  \\
                                       & Motion Vectors      &       & 1     & 1     & 3     &       &       &       &       &       &       & 1     &       &       &       &       & 1     &       & 1     & 3     & 1     &  \\
                                       & Temporal Templates  &       & 1     &       &       &       &       &       &       & 1     &       &       &       &       &       &       &       &       &       &       &       &  \\
\midrule
\multirow{2}[2]{*}{Glob. Scene}        & CRFH                &       & 1     &       &       &       &       &       &       &       &       &       &       &       &       &       & 1     &       &       &       &       &  \\
                                       & GIST                & 1     & 2     &       &       &       &       &       &       & 1     &       &       &       &       &       &       & 2     &       &       &       & 1     &  \\ 
\midrule
\multirow{2}[2]{*}{Img. Segment.}      & Superpixels         & 2     & 2     & 1     &       &       &       &       & 2     & 3     &       &       &       &       &       &       &       &       &       &       &       &  \\
                                       & Blobs               & 2     &       &       &       &       &       &       &       & 1     & 1     &       &       &       &       &       &       &       &       &       &       &  \\
\midrule
Contour                                & OWT-UCM             & 1     & 3     &       &       &       &       & 2     & 2     &       &       &       &       &       &       &       &       &       &       &       &       &  \\
\midrule
Color                                  & Histograms          & 21    & 20    & 11    & 10    &       &       & 3     & 8     & 20    & 4     & 4     & 1     & 5     &       &       & 7     &       & 1     &       & 2     &  \\
\midrule
Shapes                                 & HOG                 & 6     & 4     &       & 3     &       & 1     &       & 2     & 5     & 1     & 1     &       & 3     &       &       & 1     &       &       & 1     &       &  \\
\midrule
Orientation                            & Gabor               &       &       &       & 1     &       &       &       &       &       &       &       &       & 1     &       &       &       &       &       &       &       &  \\
\arrayrulecolor{black}
\bottomrule
\end{tabular}\label{tab:subtvf}%
\end{table*}

\begin{tabular}{rrrrrrrrrrrrrrrrrrrrrrr}
\toprule
\bottomrule
\end{tabular}%

From our previous studies in \emph{Hand-detection} and \emph{Hand-segmentation}
using multiple features and \emph{superpixels}, we want to point out that
\emph{Color} features are a good approach, particularly if a suitable color
space is exploited \cite{Morerio2013}. We found that low level features such as
\emph{Color Histograms} could help to reduce the computational complexity of
the methods and get close to real time applications. On the other side, under
large illumination changes, in \cite{Betancourt2014a} we highlight how
\emph{Color-based} hand-segmentators could introduce and disseminate in the
system noise created by hands missdetections. To alleviate this problem, we
used shape features, such as HOG, in order to pre-filter wrong measurements and
improve the classification rate of the overall system.

The two empty columns in table \ref{tab:subtvf} can be explained as follows:
\emph{Activity as a sequence} is usually chained with the output of a short
activity identification \cite{Aoki1999, Yi2009, Aghazadeh2011}, whereas
identification of the \emph{User Posture} is accomplished in \cite{Blum2006}
without employing visual features, but using GPS and accelerometers.

\subsection{Methods and algorithms} \label{sec:methalg}

Once that features are selected and estimated, the next step is to use them as
inputs to reach the objective (outputs). At this point, quantitative methods
start playing the main role, and as expected, an appropriate selection directly
influences the quality of the results, ultimately showing whether the
advantages of the FPV perspective are being exploited, or whether the
FPV-related challenges are impacting the objectives negatively. Table
\ref{tab:algorithms} shows the number of occurrences of each method (rows)
being used to accomplish a particular objective or a subtask (columns). 

\begin{table*}
    \caption{Mathematical and computational methods used in objective or each subtask}
  \centering
    \begin{threeparttable}[b]
    \begin{tabular}{r|rrrrrr|rrrrrrrrrrrrrrrr}
\toprule
      & \multicolumn{6}{c|}{Objective}                 & \multicolumn{15}{c}{Subtasks} \\
\midrule
      & \begin{sideways}Object Recognition and Tracking\end{sideways} & \begin{sideways}Activity Recognition\end{sideways} & \begin{sideways}User Machine interaction\end{sideways} & \begin{sideways}Video Summarization and Retrieval\end{sideways} & \begin{sideways}Environment Mapping\end{sideways} & \begin{sideways}Interaction Detection\end{sideways} & \begin{sideways}Background Substraction\end{sideways}& \begin{sideways}Object Identification\end{sideways} & \begin{sideways}Hand Detection and segementation\end{sideways} & \begin{sideways}People detection\end{sideways} & \begin{sideways}Object Tracking\end{sideways} & \begin{sideways}Gesture Clasification\end{sideways} & \begin{sideways}Activity Identification\end{sideways} & \begin{sideways}Activity as Sequence Analysis\end{sideways} & \begin{sideways}User Posture Detection\end{sideways} & \begin{sideways}Global Scene Identification\end{sideways} & \begin{sideways}2D-3D Secene Mapping\end{sideways} & \begin{sideways}User Personal Interests\end{sideways} & \begin{sideways}Head Movement\end{sideways} & \begin{sideways}Gaze estimation\end{sideways} & \begin{sideways}Feedback Location\end{sideways} \\ \midrule
3D Mapping          & 3     & 1     &       &       & 4     &       &       &       &       &       &       &       &       &       &       &       & 5     &       &       &       &  \\
Classifiers         & 21    & 29    & 3     & 9     & 1     & 2     & 3     & 17    & 15    & 2     &       & 2     & 15    &       &       & 4     &       & 1     & 2     & 1     &  \\
Clustering          & 4     & 8     & 3     & 5     &       & 2     &       & 3     & 6     &       &       &       & 3     &       &       & 8     &       &       &       & 1     &  \\
Comon sense         & 3     & 8     &       &       &       & 1     &       & 3     & 3     &       &       &       & 2     &       &       &       & 1     &       &       & 3     &  \\
DPMM                &       & 1     &       & 2     &       &       &       &       &       &       &       &       & 3     &       &       &       &       &       &       &       &  \\
Feature Encoding    & 4     & 6     &       & 3     &       &       &       & 6     &       &       &       &       & 3     &       &       & 3     &       &       &       &       &  \\
Optimization        &       & 1     & 1     & 1     &       &       &       &       &       &       &       &       & 1     &       &       &       &       &       &       &       & 1 \\
PGM                 & 6     & 17    &       & 3     &       & 3     & 2     & 1     &       &       &       & 1     & 11    & 1     &       & 6     &       &       & 1     & 7     &  \\
Pyramid Search      & 4     & 4     &       &       &       & 1     &       & 3     & 2     & 2     &       &       & 2     &       &       &       &       &       &       &       &  \\
Regresions          & 1     & 1     &       &       & 1     &       &       &       &       & 1     &       &       &       &       &       & 1     & 1     &       & 1     & 1     &  \\
Temporal Alignament &       &       &       & 1     &       &       &       &       &       &       &       &       &       & 1     &       &       &       &       &       &       &  \\
Tracking            & 4     &       & 3     &       & 5     &       &       &       &       &       & 7     &       &       &       &       &       & 1     &       & 2     &       &  \\
\bottomrule
    \end{tabular}\label{tab:algorithms}%
      \begin{tablenotes}
          \item [PGM] Probabilistic Graphical Models.
          \item [DPMM] Dirichlet Process Mixture Models.
          \end{tablenotes}
\end{threeparttable}
\end{table*}%
\begin{tabular}{rrrrrrrrrrrrrrrrrrrrr}

\end{tabular}%

The table highlights \emph{classifiers} as the most popular tool in FPV, which
is commonly used to assign a category to an array of characteristics (see
\textit{\cite{Lu2007}} for a more detailed survey on classifiers). The use of
classifiers is wide and varies from general applications, such as scene
recognition \cite{Fathi2011a}, to more specific, such as activity recognition
given a set of objects \cite{Fathi2012a}. Particularly, we found that the most
used are the Support Vector Machines (SVM) due to their capability to deal with
non-separable non-linear multi-label problems using low computational
resources. On the other hand, SVMs require large labeled training sets which
restricts the range of potential applications. 

In our previous works we performed a comparison of the performance of multiple
features (HOG, GIST, Color) and classifiers (SVM, Random Forest, Random Threes)
to solve the hand-detection problem \cite{Betancourt2014a}. Our conclusion was
that HOG-SVM was the best performing combination, achieving a classification
rate of $90\%$ and $93\%$ of true positives and true negatives respectively.
Another group of methods commonly used are \emph{clustering} algorithms due to
its simplicity, computational cost, and small requirements in the training
datasets. Despite their advantages, \emph{clustering} algorithms could require
post-processing analysis of the results in order to endow them with human
interpretation.

Another promising group of tools are the \emph{Probabilistic Graphical Models}
(PGMs), which can be interpreted as a framework to combine multiple sensors and
chain results from different methods in a unique probabilistic hierarchical
structure (e.g. to recognize the object and subsequently use it to infer the
activity).  \emph{Dynamic Bayesian Networks} (DBNs) are a particular type of
PGMs which include time in their structure, in turn making them suitable for
application in video analysis \textit{\cite{crowdchapter}}. As an example, DBNs
are frequently used to represent activities as sequences of events
\cite{Starner1998, Schiele1999a,Clarkson2000, Blum2006, Sundaram2009}.
{\color{black}It is common to find that particular methods, such as Dirichlet
Process Mixture Models (DPMM), are presented in their PGM notation, however
given the promising recent results achieved in \emph{Activity
Recognition} and \emph{Video Segmentation}, we decided to group them
separately.}


As stated in section \ref{sec:features}, there is a large number of features
that can be extracted for FPV applications. A common practice is to mix or
chain multiple features before using them as input of a particular algorithm
(table \ref{tab:algorithms}). This practice usually results in extremely large
vectors of features that can lead to computationally expensive algorithms. In
this context, the role of \emph{Feature Encoding} methods, such as
Bag-of-Words, is crucial to control the size of the inputs. We highlight the
importance that some authors are giving to this tool, which, despite not being
an automatic strategy like Linear Discriminant Analysis (LDA) and Principal
Components Analysis (PCA), can nevertheless help to include human intuition in
the analysis. As an example, the authors in \cite{Matsuo2014} use BoW in
\emph{Activity Recognition} taking into account the presence, level of
attention, and the role of the objects in the video. 

The use of machine learning methods (e.g. classifiers, clustering, regressors)
introduces an important question to the analysis: how to train the algorithms
on realistic data without restricting their applicability? This question is
widely studied in the field of Artificial Intelligence, and two different
approaches are commonly followed, namely unsupervised and supervised learning
\textit{\cite{Camastra2007}}. Unsupervised learning requires less human
interaction in training steps, but requires human interpretation of the
results. Additionally, unsupervised methods have the advantage of being easily
adaptable to changes in the video (e.g. new objects in the scene or
uncontrolled environments \cite{Spriggs2009}). The most commonly used
unsupervised method in FPV are the clustering algorithms, such as k-means. In
fact, the best performing superpixels are the result of an unsupervised
clustering procedure applied over a raw image\textit{\cite{Slic_2012}}.  In
\cite{morerio2014optimizing} we proposed an optimization of the SLIC
superpixels, and latter in \textit{\cite{morerio2014generative}} we introduced
a new superpixel method based on Neural Networks. The proposed algorithm is a
self-growing map that adapts its topology to the frame structure taking
advantage of the dynamic information available in the previous frames.  

Regarding the supervised methods, their results are easily interpretable but
commonly imply higher requirements in the training stage. As an example, at the
beginning of this section we highlighted some of the applications of SVMs.
Supervised methods use a set of inputs, previously labeled, to parametrize the
models. Once the method is trained, it can be used on new instances without any
additional human supervision. In general, supervised methods are more dependent
on the training data, fact which could work against their performance when used
on newly-introduced cases \cite{Pirsiavash2012, Lu2013, Spriggs2009, Fathi2012,
Wu2007, Ghosh2012, Lee2015}. In order to reduce the training requirements, and
take advantage of the useful information available on Internet, some authors
create their datasets using services like Amazon Mechanical Turk
\cite{Spain2010, Ghosh2012}, automatic web mining \cite{Berg2006, Wu2007}, or
image repositories \cite{Xiong2014}. We named this practice in table
\ref{tab:algorithms} as \emph{Common Sense}.

\emph{Weakly supervised learning} is another commonly used strategy, considered
as a middle point between supervised and unsupervised learning. This strategy
is used to improve the supervised methods in two aspects: i) extending the
capability of the method to deal with unexpected data; and ii) reducing the
necessity for large training datasets.  Following this trend, the authors of
\cite{Dovgalecs2010,Karaman2010} used Bag of Features (BoF) to monitor the
activity of people with dementia. Later, \cite{Fathi2011a,Fathi2011} used
Multiple Instance Learning (MIL) to recognize objects using general categories.
Afterwards, \cite{Aghazadeh2011} used BoF and Vector of Locally Aggregated
Descriptors (VLAD) to temporally align a sequence of videos. Eventually, let us
mention \emph{Deep learning}, a relatively recent approach which combines
supervised and unsupervised learning techniques in a unified framework, where
low level significant features are learned in an unsupervised fashion
\textit{\cite{deeplearning2012}}.

\section{Public datasets} \label{sec:datasets}

In order to support their results and create benchmarks in FPV video analysis,
some authors have provided their datasets for public use to the academic
community. The first publicly available FPV dataset is released by
\cite{Mayol2005}. It consists of a video containing 600 frames recorded in a
controlled office environment using a camera on the left shoulder, while the
user interacts with five different objects. Later, \citep{Philipose2009}
proposed a larger dataset with two people interacting with 42 object instances.
The latter one is commonly considered as the first challenging FPV dataset
because it guaranteed the requirements identified by \cite{Schiele1999}: i)
Scale and texture variations, ii) Frame resolution, iii) Motion blur, and iv)
Occlusion by hand. 

\begin{table*}
  \caption{Current datasets and sensors availability}
  \centering
\begin{threeparttable}[b]
\begin{tabular}{rr|l|c|l|c|cccccc|ccc|ccc}
\toprule
\multicolumn{6}{c}{} & \multicolumn{6}{c}{Sensors}           & \multicolumn{3}{c}{\# Objects} & \multicolumn{3}{c}{Cam. Location} \\
\midrule
\multicolumn{1}{c}{} & \multicolumn{1}{c|}{\begin{sideways}\end{sideways}} & \multicolumn{1}{c|}{\begin{sideways}Year\end{sideways}} & \multicolumn{1}{c|}{\begin{sideways}Location\end{sideways}} & \multicolumn{1}{c|}{\begin{sideways}Controlled Conditions\end{sideways}} & \multicolumn{1}{c|}{\begin{sideways}Objective\end{sideways}} & \begin{sideways}Video\end{sideways} & \begin{sideways}Depth-Sensor\end{sideways} &\begin{sideways}IMUs\end{sideways} & \begin{sideways}Body Media\end{sideways} & \begin{sideways}eWatch\end{sideways} & \begin{sideways}Eye Tracking\end{sideways} & \begin{sideways}Activities\end{sideways} & \begin{sideways}Objects\end{sideways} & \begin{sideways}Num. of people\end{sideways} & \begin{sideways}Shoulder\end{sideways} & \begin{sideways}Chest\end{sideways} & \begin{sideways}Head\end{sideways} \\ \midrule
\href{http://www.youtube.com/watch?v=P6pXgDlB5a0}{Mayol05 } & \cite{Mayol2005}        & 2005  & Desktop & \ding{51} & O1    & \ding{51} &  &       &       &       &       &       & 5     & 1     & \ding{51}  &       &  \\
\href{http://seattle.intel-research.net/~xren/egovision09/}{Intel } & \cite{Philipose2009}    & 2009  & Multiple locations &       & O1    & \ding{51} &  &       &       &       &       &       & 42    & 2     & \ding{51}  &       &  \\
\href{http://kitchen.cs.cmu.edu/}{Kitchen. } & \cite{Spriggs2009}      & 2009  & Kitchen Recipes & \ding{51} & O2    & \ding{51} &  & \ding{51} & \ding{51} & \ding{51} &       & 3     &       & 18    &       &       & \ding{51}  \\
\href{http://ai.stanford.edu/~alireza/GTEA/}{GTEA11 } & \cite{Fathi2011}        & 2011  & Kitchen Recipes & \ding{51} & O2    & \ding{51} &  &       &       &       &       & 7     &       & 4     &       &       & \ding{51}  \\
\href{http://www.csc.kth.se/cvap/vinst/NovEgoMotion.html}{VINST } & \cite{Aghazadeh2011}    & 2011  & Going to the work &       & O2    & \ding{51} &  &       &       &       &       &       &       & 1     &       & \ding{51}  &  \\
\href{http://www.cs.cmu.edu/~kkitani/EgoAction.html}{UEC Dataset } & \cite{Kitani2011}       & 2011  & Park  &       & O2    & \ding{51} &  &       &       &       &       & 29    &       & 1     &       &       & \ding{51}  \\
\href{http://people.csail.mit.edu/hpirsiav/codes/ADLdataset/adl.html}{ADL } & \cite{Pirsiavash2012}   & 2012  & Daily activities &       & O2    & \ding{51} &  &       &       &       &       & 18    &       & 20    &       & \ding{51}  &  \\
\href{http://vision.cs.utexas.edu/projects/egocentric/index.html}{UTE } & \cite{Ghosh2012}        & 2012  & Daily activities &       & O4    & \ding{51} &  &       &       &       &       &       &       & 4     &       &       & \ding{51}  \\
\href{http://ai.stanford.edu/~alireza/Disney/}{Disney } & \cite{Fathi2012}        & 2012  & Thematic Park &       & O6    & \ding{51} &  &       &       &       &       &       &       & 8     &       &       & \ding{51}  \\
\href{http://ai.stanford.edu/~alireza/GTEA_Gaze_Website/GTEA_Gaze+.html}{GTEA gaze} & \cite{Fathi2012a}       & 2012  & Kitchen Recipes & \ding{51} & O2    & \ding{51} &  &       &       &       & \ding{51} & 7     &       & 10    &       &       & \ding{51}  \\
\href{http://www.cs.cmu.edu/~kkitani/perpix/}{EDSH } & \cite{Li2013a}          & 2013  & Multiple locations &       & O1    & \ding{51} &  &       &       &       &       & -     & -     & -     &       &       & \ding{51}  \\
\href{http://cvrc.ece.utexas.edu/mryoo/jpl-interaction.html}{JPL } & \cite{Ryoo2013}         & 2013  & Office Building &       & O6    & \ding{51} &  &       &       &       &       & 7     &       & 1     &       &       & \ding{51}  \\
\href{http://imagelab.ing.unimore.it/files/EGO-HSGR.zip}{EGO-HSGR } & \cite{Serra2013}        & 2013  & Library Exhibition &       & O3    & \ding{51} &  &       &       &       &       & 5     &       & 1     &       &       & \ding{51}  \\
\href{http://www.cs.bris.ac.uk/~damen/BEOID/}{BEOID } & \cite{Damen2014} & 2014  & Multiple locations &       & O2    & \ding{51} &  &       &       &       & \ding{51} & 6     &       & 5     &       &       & \ding{51}  \\
\href{http://imagelab.ing.unimore.it/files/EGO-GROUP.zip}{EGO-GROUP } & \cite{Alletto2014} & 2014  & Multiple locations &       & O6    & \ding{51} &  &       &       &       &       &       &       & 19    &       &       & \ding{51}  \\
\href{http://imagelab.ing.unimore.it/files/EGO-HPE.zip}{EGO-HPE } & \cite{Alletto} & 2014  & Multiple locations &       & O1    & \ding{51} &  &       &       &       &       &       &       & 4     &       &       & \ding{51}  \\
\href{http://www.vision.huji.ac.il/egoseg/videos/dataset.html}{EgoSeg} & \cite{Poleg2014} & 2014  & Multiple locations &       & O2    & \ding{51} &   &       &       &       &       &   7   &       & 2     &       &       & \ding{51}  \\
\href{http://www.ics.uci.edu/~jsupanci/HANDS-2015/#challange}{Egocentric Intel/Creative} & \cite{Rogez2014} & 2014  & Multiple locations &       & O1    & \ding{51} &   \ding{51}    &       &       &       &      &       &   2   &       &       &  \ding{51} \\
\bottomrule
\end{tabular}%
  \begin{tablenotes}
      \item [*] {\bf Objectives: } [O1] Object Recognition and Tracking. [O2] Activity Recognition. [O3] User-Machine Interaction. [O4] Video Summarization. [O5] Phisical Scene Reconstruction. [O6] Interaction Detection.
      \item [**] The table summarizes the characteristic described in the technical reports or the papers proposing the datasets.
      \end{tablenotes}
\end{threeparttable}
 \label{tab:datasets2}
\end{table*}%

%
Implicitly, previous sections explain some of the main characteristics of FPV
videos. In \cite{Tan}, these characteristics are compared for several FPV and
Third Person Vision (TPV) datasets and their classification capabilities are
evaluated. The authors reach a classification accuracy of $80.9\%$ using blur,
illumination changes, and optical flow as input features. In their study they
also found a considerable difference in the classification rate explained by
the camera position.  The authors concluded that the more stable the camera,
the less blur and motion and then the less discriminative power of these
features. We highlight this difference as an important finding because it opens
the door to an interesting discussion concerning which kind of videos, based on
quantitative measurements, should be considered as FPV. Extra evidence about
the role of the non-wearable cameras, such as hand-held devices when they are
used to record from a first person perspective, is still pending. Our intuition
points that, despite having some of the challenging characteristics of wearable
cameras like mobile backgrounds and unstable motion patterns, hand-held videos
would drastically differ in terms of features compared in \cite{Tan}.

Table \ref{tab:datasets2} presents a list of the publicly-available datasets,
along with their characteristics. Of particular interest are the changes in the
camera location, which have evolved from shoulder-based to the head-based.
These changes are clearly explained by the trend of the smart-glasses and
action cameras (see Table \ref{tab:commercial}). Also noticeable are the
changes in the objectives of the datasets, moving from low level, such as
object recognition, to more complex objectives, such as social interaction and
user-machine interaction. It should also be noted that less controlled
environments have recently been proposed to improve the robustness of the
methods in realistic situations. In order to highlight the robustness of their
methods, several authors evaluated them on Youtube sequences recorded using
goPro cameras \cite{Kitani2011}.

Another aspect to highlight from the table is the availability of multiple
sensors in some of the datasets. For instance, the Kitchen dataset
\cite{Spriggs2009} includes four sensors, the GTEA approach \cite{Fathi2012a}
includes eye tracking measurements, and the Egocentric Intel/Creative
\cite{Rogez2014} was recorded with a RGBD camera. 

\section{Conclusion and future research} \label{sec:future}

Wearable devices such as smart-glasses will presumably constitute a
significant share of the technology market during the coming years, bringing
new challenges and opportunities in video analytics. The interest in the
academic world has been growing in order to satisfy the methodological
requirements of this emerging technology. This survey provides a summary of the
state of the art from the academic and commercial point of view, and summarizes
the hierarchical structure of the existent methods. This paper shows the large
number of developments in the field during the last 20 years, highlighting main
achievements and some of the up-coming lines of study.

From the commercial and regulatory point of view, important issues must be
faced before the proper commercialization of this new technology can take
place. Nowadays, the privacy of the recorded people is one of the most
discussed ones, as these kinds of devices are commonly perceived as intruders
\cite{Nguyen2009}. Other important aspects are the legal regulations depending
on the country, {\color{black}, and the intention of the user to avoid
recording private places or activities\cite{Templeman2014}. Another hot topic
is the real applicability of smart-glasses as a massive consumption device or
as a task-oriented tool to be worn only in particular scenarios.  In this
field, the technological companies are designing their strategies in order to
reach out to specific markets. As an illustration, recent turn of events has
seen Google move out of the glass project (originally intended to end with a
massively commercialized product), in order to target the enterprise market.
Microsoft, on the other hand, recently announced  its task-oriented
holographic device ``HoloLens'' embodied with a larger array of sensors.}

From the academic point of view, the research opportunities in FPV are still
wide. Under the light of this bibliographic review and our personal experience,
we identify 4 main hot topics:

\begin{itemize}

\item{Existing methods are proposed and executed in previously recorded videos.
However, none of them seems to be able to work in a \emph{closed-loop} fashion,
by continuously learning from users' experiences and adapt to the highly
variable and uncontrollable surrounding environment. From our previous studies
\textit{\cite{FUSION2013, JournalAIHC2014}}, we believe that a cognitive
perspective could give important cues to this aspect and could aid the
development of the self-adaptive devices.} 

\item{The \emph{personalization} capabilities of smart-glasses open the door to
new learning strategies. Incoming methods should be able to receive
personalized training from the owner of the device. We have found out, for
instance, that this kind of approach can help alleviate problems, such as
changes in the color skin models from different users \cite{Morerio2013} in a
hand detection application. Indeed, color features, as stressed in
\ref{tab:subtvf}, has proven to be extremely suitable to be exploited in this
field.} 

\item{This survey focuses on methods for addressing tasks accomplished mainly
by one user coupled with a single wearable device. However, \emph{cooperative
devices} would be useful to increase the number of applications in areas such
as environment mapping, military applications, cooperative games, sports, etc.}

\item{Finally, regarding the \emph{real time requirements}, important
developments should be made in order to optimally compute FPV methods without
draining the battery. This must be accomplished both from the hardware and the
software side. On the one hand, progress still needs to be made on the
processing units of the devices. On the other, lighter, faster and better
optimized methods are yet to be designed and tested. Our personal experience
lead us to explore fast machine learning methods \cite{Betancourt2014a} for
hand detection, in the trend highlighted by table \ref{tab:algorithms}, and to
discard standard features such as optic flow \cite{Morerio2013} because of
computational restrictions. Promising methods in standard computer vision
research, such as superpixel methods, were built from scratch in
\textit{\cite{morerio2014generative}} in order to make them faster and better
suited for video analysis \cite{morerio2014optimizing}. Eventually, important
cues to the problem of computational power optimization may also be found in
cloud computing and high performance computing.} 
\end{itemize}

\scriptsize
\bibliographystyle{ieeetr}
\bibliography{references,biblio}


\begin{IEEEbiography}[{\includegraphics[width=1in,height=1.5in,clip,keepaspectratio]{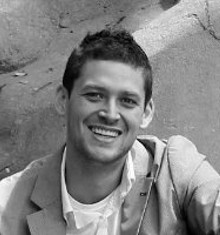}}]{Alejandro Betancourt}
Alejandro Betancourt is PhD candidate of the Interactive and Cognitive Environment program between the Universita degli Studi di Genova and the Eindhoven University of Technology.   Alejandro is a Mathematical Engineer and Master In Applied Mathematics from EAFIT University (Medellin, Colombia). Since 2011 Alejandro has been involved in research about Artificial Intelligence, Machine Learning and Cognitive Systems.
\end{IEEEbiography}
\begin{IEEEbiography}[{\includegraphics[width=1in,height=1.5in,clip,keepaspectratio]{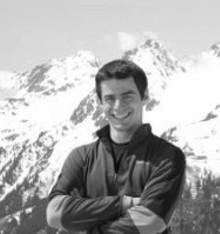}}]{Pietro Morerio}
Pietro Morerio received his B.SC in Physics from the Faculty of Science, University of Milan (Italy) in 2007. In 2010 he received from the same university his M. Sc. in Theoretical Physics (summa cum laude). He was Research Fellow at the University of Genoa (Italy) from 2011 to 2012, working in Video Analysis for Interactive Cognitive Environments. Currently, he is pursuing a PhD degree in Computational Intelligence at the same institution.
\end{IEEEbiography}
\begin{IEEEbiography}[{\includegraphics[width=1in,height=1.5in,clip,keepaspectratio]{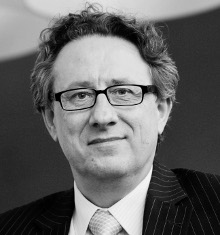}}]{Matthias Rauterberg}
Matthias Rauterberg is professor at the department of Industrial Design and the head of the Designed Intelligence group at Eindhoven University of Technology (The Netherlands). Matthias received the B.S. in Psychology (1978) at the University of Marburg (Germany), the B.S. in Philosophy (1981) and Computer Science (1983), the M.S. in Psychology (1981, summa cum laude) and Computer Science (1986, summa cum laude) at the University of Hamburg (Germany), and the Ph.D. in Computer Science/Mathematics (1995, awarded) at the University of Zurich (Switzerland).
\end{IEEEbiography}
\begin{IEEEbiography}[{\includegraphics[width=1in,height=1.55in,clip,keepaspectratio]{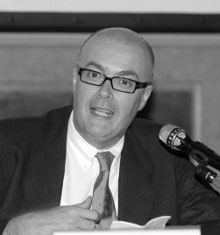}}]{Carlo Regazzoni}
Carlo S. Regazzoni received the Laurea degree in Electronic Engineering and the Ph.D. in Telecommunications and Signal Processing from the University of Genoa (UniGE), in 1987 and 1992, respectively. Since 2005 Carlo is Full Professor of Telecommunications Systems. Dr. Regazzoni is involved in research on Signal and Video processing and Data Fusion in Cognitive Telecommunication Systems since 1988.
\end{IEEEbiography}

\end{document}